\documentclass{article}

\usepackage[utf8]{inputenc} 
\usepackage[T1]{fontenc}    
\usepackage{booktabs}       
\usepackage{nicefrac}       
\usepackage{microtype}      
\usepackage{times}             

\usepackage[round]{natbib}
\usepackage{amssymb,amsmath,amsthm,bbm}
\usepackage[margin=1in]{geometry}
\usepackage{verbatim,float,url,dsfont}
\usepackage{graphicx,subfigure,psfrag}
\usepackage{algorithm,algorithmic}
\usepackage{mathtools,enumitem}
\usepackage[colorlinks,citecolor=blue]{hyperref}
\usepackage[utf8]{inputenc} 
\usepackage[T1]{fontenc}    
\usepackage{hyperref}       
\usepackage{url}            
\usepackage{booktabs}       
\usepackage{amsfonts}       
\usepackage{nicefrac}       
\usepackage{microtype}      

\usepackage{times}             
\usepackage{amsthm}

\usepackage{bbm}
\usepackage{verbatim,float,dsfont}
\usepackage{psfrag}
\usepackage{algorithm,algorithmic}
\usepackage{mathtools,enumitem}
\mathtoolsset{showonlyrefs}
\usepackage{tcolorbox}
\usepackage{breqn,lipsum}
\usepackage{thmtools,thm-restate}
\usepackage{graphicx,stackengine}
\usepackage{array}
\usepackage{caption}
\usepackage{boldline}
\usepackage{amssymb}
\usepackage{multirow}

\newtheorem{theorem}{Theorem}
\newtheorem{lemma}[theorem]{Lemma}
\newtheorem{corollary}[theorem]{Corollary}
\newtheorem{proposition}[theorem]{Proposition}
\newtheorem{remark}[theorem]{Remark}
\newtheorem{definition}[theorem]{Definition}

\DeclarePairedDelimiter\floor{\lfloor}{\rfloor}

\def\shownotes{1}  
\ifnum\shownotes=1
\newcommand{\authnote}[2]{$\ll$\textsf{\footnotesize #1 notes: #2}$\gg$}
\else
\newcommand{\authnote}[2]{}
\fi

\newcommand{\yw}[1]{\textcolor{red}{\textbf{[yuxiang: #1]}}}

\usepackage{accents}
\newcommand{\ubar}[1]{\underaccent{\bar}{#1}}

\makeatletter
\newcommand*\rel@kern[1]{\kern#1\dimexpr\macc@kerna}
\newcommand*\widebar[1]{%
  \begingroup
  \def\mathaccent##1##2{%
    \rel@kern{0.8}%
    \overline{\rel@kern{-0.8}\macc@nucleus\rel@kern{0.2}}%
    \rel@kern{-0.2}%
  }%
  \macc@depth\@ne
  \let\math@bgroup\@empty \let\math@egroup\macc@set@skewchar
  \mathsurround\z@ \frozen@everymath{\mathgroup\macc@group\relax}%
  \macc@set@skewchar\relax
  \let\mathaccentV\macc@nested@a
  \macc@nested@a\relax111{#1}%
  \endgroup
}
\makeatother

\newcommand{\ALG}{\textsc{Aligator}}

\def\R{\mathbb{R}}

\def\Var{\mathrm{Var}}

\def\btheta{\boldsymbol{\theta}}

\def\cA{\mathcal{A}}

\def\cE{\mathcal{E}}
\def\cF{\mathcal{F}}

\def\cI{\mathcal{I}}
\def\cL{\mathcal{L}}

\def\cO{\mathcal{O}}

\def\cS{\mathcal{S}}
\def\cT{\mathcal{T}}

\def\cX{\mathcal{X}}

\def\TV{\mathrm{TV}}

\def\bs{\ensuremath\boldsymbol}

\newcommand{\red}[1]{\textcolor{red}{#1}}

\makeatletter
\newcommand*\fourteenpt{\fontsize{16}{15.5}\selectfont}
\newcommand*\totht[1]{\dimexpr\ht#1+\dp#1\relax}
\newcommand*\leading{{\setbox0\hbox{\strut}\the\totht0}}
\newcommand*\fntsize{{\setbox0\hbox{Mg}\the\totht0}}
\newcommand*\showsize[1]{{#1 {\ttfamily\string#1} (\f@size pt) \fntsize/\leading}\par}
\makeatother

\twocolumn

\title{\fourteenpt An Optimal Reduction of TV-Denoising to Adaptive Online Learning}  

\author{Dheeraj Baby \\dheeraj@ucsb.edu \and Xuandong Zhao \\xuandongzhao@ucsb.edu \and Yu-Xiang Wang \\yuxiangw@cs.ucsb.edu}

\date{Dept. of Computer Science, UC Santa Barbara}

\begin{document}

\maketitle

%

%

\begin{abstract}
We consider the problem of estimating a function from $n$ noisy samples whose discrete Total Variation (TV) is bounded by $C_n$. We reveal a deep connection to the seemingly disparate problem of \emph{Strongly Adaptive} online learning \citep{daniely2015strongly} and provide an $O(n \log n)$ time algorithm that attains the near minimax optimal rate of $\tilde O (n^{1/3}C_n^{2/3})$ under squared error loss. The resulting algorithm runs online and optimally \emph{adapts} to the \emph{unknown} smoothness parameter $C_n$. This leads to a new and more versatile alternative to wavelets-based methods for (1) adaptively estimating TV bounded functions; (2) online forecasting of TV bounded trends in time series.
\end{abstract}

\section{Introduction} \label{sec:intro}

\emph{Total variation} (TV) denoising \citep{tv} is a classical algorithm originated in the signal processing community which removes noise from a noisy signal $y$ by solving the following regularized optimization problem
\begin{align}\label{eq:ROF_model}
    \min_{f} \|f - y\|_2^2 + \lambda \TV(f).
\end{align}
where $\TV(\cdot)$ denotes the total variation functional which is equivalent to $\int |f'(x)| dx$ for weakly differentiable functions.  In discrete time, TV denoising is known as ``fused lasso'' in the statistics literature \citep{fuse,hoefling2010path}, which solves
\begin{align}\label{eq:fused_penalty}
    \min_{\btheta\in\R^n} \sum_{i=1}^n(\theta_i - y_i)^2 + \lambda \sum_{i=2}^n |\theta_i - \theta_{i-1}|.
\end{align}
where $\theta_i$ is the element at index $i$ of the vector $\btheta$. Unlike their L2-counterpart, the TV regularization functional is designed to promote sparsity in the number of change points, hence inducing a ``piecewise constant'' structure in the solution. 

Over the three decades since the advent of TV denoising, it has seen many influential applications. Algorithms that use TV-regularization has been deployed in every cellphone, digital camera and medical imaging devices.  More recently, TV denoising is recognized as a pivotal component  in generating the first image of a super massive black hole \citep{blackhole}.
Moreover, the idea of TV regularization has inspired a myriad of extensions to other tasks such as image debluring, super-resolution, inpainting, compression, rendering, stylization (we refer readers to a recent book \citep{chambolle2010introduction} and the references therein) as well as other tasks beyond the context of images such as change-point detection, semisupervised learning and graph partitioning.

In this paper, we focus on the \emph{non-parametric statistical estimation} problem behind TV-denoising which aims to estimate a function $f:[0,1]\rightarrow \R$ using observations of the following form:
\begin{align}
    y_i = f(x_i) + \epsilon_i, i \in [n] := \{1,\ldots,n\}, \label{eq:offline}
\end{align}
where $\epsilon_i$ are iid $N(0,\sigma^2)$ and the function $f$ belongs to some fixed non-parametric function class $\cF$. The exogenous variables $x_i$ belongs to some subset $\cX$ of $\mathbb{R}$. The above setup is a widely adopted one in the non-parametric regression literature \citep{tsybakov_book}. In this work, we take $\cF$ to be the Total Variation class: $\{f | \TV(f) \leq C_n\}$ or its discrete counterpart 
$$\cF(C_n) := \bigg\{ f \bigg| \sum_{t=2}^n|f(x_t) - f(x_{t-1})| \le C_n \bigg\}.$$


We are interested in finding algorithms that generate estimates $\hat{y}_t, t \in [n]$ such that the total square error
\begin{align}
    R_n(\hat{y}, f) := \sum_{t=1}^n\mathbb{E}[(\hat{y_t} - f(x_t))^2], \label{eq:regret}
\end{align}
is minimized. Throughout this paper, when we refer to \emph{rate}, we mean the growth rate of $R_n$ as a function of $n$ and $C_n$. The family $\cF(C_n)$ we consider here features a rich class of functions that exhibit spatially heterogeneous smoothness behavior. These functions can be very smoothly varying in certain regions of space, while in other regions, it can exhibit fast variations (see for eg. Fig. \ref{fig:dopplerfig}) or abrupt changes that may even be discontinuous. 
A good estimator should be able to detect such local fluctuations (which can be short lived) and adjust the amount of ``smoothing'' to apply according to the level of smoothness of the functions in each local neighborhood. 
Such estimators are referred as \emph{locally adaptive estimators} by Donoho \citep{donoho1998minimax}. 

We are interested in algorithms that achieve the minimax optimal rates for estimating functions in $\cF(C_n)$ defined as:
\begin{align*}
R_n^*(C_n) = \inf_{\{\hat{y}_t\}_{t=1}^{n}} \: \sup_{f \in \cF(C_n)} R_n(\hat{y},f),
\end{align*}
which is known to be $\Theta(n^{1/3}C_n^{2/3})$\citep{donoho1990minimax,mammen1991}.

There is a body of work in \emph{Strongly Adaptive online learning } that focuses on designing online algorithms such that its regret in any local time window is controlled \citep{daniely2015strongly}. Hence the notion of local adaptivity is built into such algorithms. This makes the problem of estimating TV bounded functions, a natural candidate to be amenable to techniques from Strongly Adaptive online learning. However, it is not clear that whether using Strongly Adaptive algorithms can lead to minimax optimal estimation rates. By formalizing the intuition above, we answer it affirmatively in this work.

We reserve the phrase \emph{adaptive estimation} to describe the act of estimating TV bounded functions such that  $R_n$ of the estimator/algorithm can be bounded by a function of $n$ and $C_n$ without any prior knowledge of $C_n$.  An \emph{adaptively optimal} estimator $\hat{y}$ is able to estimate an arbitrary function $f$ with an error 
$$ R_n(\hat{y},f) = \tilde{O}\Big(\inf_{C_n \text{ such that } f\in \cF(C_n) }R_n^*(C_n) \Big). $$
A TV bounded function will be referred as a Bounded Variation (BV) function henceforth for brevity.The notation $\tilde O (\cdot)$ hides poly-logarithmic factors of $n$.

It is well known that \emph{all linear estimators} that output a linear transformation of the observations attain a suboptimal $\Omega(\sqrt{nC_n})$ rate \citep{donoho1990minimax}. This covers a large family of algorithms including the popular methods based on smoothing kernels, splines and local polynomials, as well as methods such as online gradient descent \citep[see a recent discussion from ][]{arrows2019}. Wavelet smoothing \citep{donoho1998minimax} is known to attain the near minimax optimal rate of $\tilde O(n^{1/3}C_n^{2/3})$ for $R_n$ without any prior information about $C_n$.  Recently the same rate is shown to be achievable for the online forecasting setting by adding a wavelets-based adaptive restarting schedule to OGD \citep{arrows2019}.

In this paper, we provide an alternative to wavelet smoothing by a novel reduction to a strongly adaptive regret minimization problem from the online learning literature. 
We show that the resulting algorithm achieves the same adaptive optimal rate of $\tilde O(n^{1/3}C_n^{2/3})$. The algorithm is more versatile than wavelet smoothing for three reasons:
\begin{enumerate}
\item Our algorithm is based on aggregating experts that performs local predictions. The experts we use perform online averaging. However, one may use more advanced algorithms such as kernel/spline smoothing, polynomial regression or even deep learning approaches as experts that can potentially lead to better performance in practice. Hence our algorithm is highly configurable.

\item Our algorithm accepts a learning rate parameter that can be set without prior knowledge of $C_n$ to obtain the near optimal rate of $\tilde O(n^{1/3}C_n^{2/3})$ (see Theorem \ref{thm:main}). However, this learning rate can also be tuned using heuristics that can lead to better practical performance (see Section \ref{sec:exp}).

\item It can also handle a more challenging setting where the data are streamed sequentially in an online fashion.
\end{enumerate}

To the best of our knowledge, we are the first to formalize the connection between strongly adaptive online learning and the problem of local-adaptivity in nonparametric regression. By establishing this new perspective, we hope to encourage further collaboration between these two communities.

\subsection{Problem Setup}
\label{sec:problem_setup}

\begin{figure}[t]
	\centering
	\fbox{
		\begin{minipage}{7 cm}
			\begin{enumerate}
				\setlength\itemsep{0em}
				\item Player (we) declares a forecasting strategy
				\item Adversary chooses an $\cX = \{x_1 < x_2 < \ldots < x_n \}$ and reveals it to the player.
				\item Adversary chooses $f(x_1),\ldots,f(x_n)$ such that $\sum_{t=2}^n |f(x_t) - f(x_{t-1})| \le C_n$.
				\item Adversary fixes an ordered set $\{i_1,\ldots, i_n \}$ where each $i_j \in [n]$.
				\item For every time point $t = 1,...,n$:
				\begin{enumerate}
					\item Adversary reveals $i_t$.
					\item We play $\hat{y}_t$.
					\item We receive a feedback\\ $y_t = f(x_{i_t}) +  \epsilon_t$,
					\\where $\epsilon_t$ is $N(0,\sigma^2)$.
					\item We suffer loss $(\hat y_t - y_t)^2$
				\end{enumerate}
				\item Our goal is to minimize\\ $\sum_{t=1}^{n}\mathbb{E}[(\hat y_{t} - f(x_{i_t}))^2]$.
			\end{enumerate}
		\end{minipage}
	}
	\caption{\emph{Online interaction protocol}}
	\label{fig:online-protocol}
\end{figure}
Though we are primarily motivated to solve the offline/batch estimation problem, our starting point is to consider a significant generalization of the batch problem as shown in Fig. \ref{fig:online-protocol}. Any adaptively optimal algorithm to this online game immediately implies adaptive optimality in the batch/offline setting. For example, to solve the batch problem, adversary can be thought of as revealing the indices isotonically, i.e $i_t = t$. However, note that in the online game, adversary can even query the same index multiple times. The term ``forecasting strategy'' in step 1 of  Fig. \ref{fig:online-protocol}, is used to mean an algorithm that makes a prediction at current time point only based on the historical data.


Solving the online problem has an added advantage that the resulting algorithm can be applied to various instances of time series forecasting like financial markets, spread of contagious disease etc.

\textbf{Assumption 1} \label{assumption} $|f(x_i)| \le B, \forall i \in [n]$ for some known $B$.

Though this constraint is considered to be mild and natural, we note that standard non-parametric regression algorithms 
do not make this assumption.

\subsection{Notes on novelty and contributions} 
To the best of our knowledge, in non-parametric regression literature, only wavelet smoothing \footnote{Though \citep{arrows2019} proposes a minimax policy for forecasting TV bounded sequences online, they heavily rely on the adaptive minimaxity of wavelet smoothing.} \citep{donoho1998minimax} is able to \emph{provably} attain a near optimal $\tilde O(n^{1/3}C_n^{2/3})$ rate for estimating BV functions in batch setting without knowing the value of $C_n$. There are model-selection techniques based on information-criterion, which often either incurs significant practical overhead or comes with no optimal rate guarantees (We will review these approaches in Section \ref{sec:lit}).

The contributions of this work is mainly theoretical. Our primary result is a novel reduction from the problem of estimating BV functions to Strongly Adaptive online learning \citep{daniely2015strongly}. This reduction approach results in the development of a new $O(n \log n)$ time algorithm that is: 1) \emph{minimax optimal} (modulo log factors) 2) \emph{adaptive} to $C_n$ and 3) can be used to tackle \emph{both} online and offline estimation problems thereby providing new insights. To elaborate slightly, this is facilitated by few fundamentally different viewpoints than those adopted in the  wavelet literature. In particular, we exhibit a specific partitioning of TV bounded function into consecutive chunks that incurs low total variation such that total number of chunks is $O(n^{1/3}C_n^{2/3})$. Then by designing a strongly adaptive online learner, we ensure an $\tilde O(1)$ cumulative squared error in each  chunk of that partition. This immediately implies an estimation error rate of $\tilde O(n^{1/3}C_n^{2/3})$ when summed across all chunks. To the best of our knowledge, this is the \emph{first} time a connection between strongly adaptive online learning and estimating BV functions has been exploited in literature.

Experimental results (see Section \ref{sec:exp}) indicate that our algorithm can outperform wavelet smoothing in terms of its cumulative squared error incurred in practice. We demonstrate that the proposed algorithm can be used without any hyper-parameter tuning and incurs very low computational overhead in comparison to model selection based approaches for the fused lasso problem (see Eq. \eqref{eq:fused_penalty}).

Before closing this section, we remind the reader that this work shouldn’t be viewed only as providing yet another solution to a classical problem  but rather one that provides \emph{a fundamentally new set of tools} that adds new insight to this decades-old problem that might have a profound impact in many extensions of the basic setting we consider and other downstream tasks such as  estimating higher-dimensional BV functions, fused lasso on graphs, image deblurring, trend filtering and so on.

\subsection{Related Work} \label{sec:lit}
As noted before, the theoretical analysis of estimating BV functions is well studied in the rich literature of non-parametric regression. Apart from wavelet smoothing \citep{donoho1990minimax,donoho1994ideal,donoho1994minimax,donoho1998minimax}, many algorithms such as Trend Filtering \citep{l1tf,tibshirani2014adaptive,graphtf,sadhanala2016graph,guntuboyina2018constrainedTF} and locally adaptive regression splines \citep{locadapt} can be used for  estimation. However, one drawback of these algorithms is that they require the TV of ground truth $C_n$ as an input to the algorithm to guarantee minimax optimal rates. For example, the solution to fused lasso (Eq. \eqref{eq:fused_penalty}) is minimax optimal only when one chooses the hyper-parameter $\lambda$ optimally. It is shown in \citep{graphtf} that optimal choice of $\lambda$ depends on the variational budget $C_n$ which may be unknown beforehand.

Theoretically one may tune the choice $C_n$ (or $\lambda$) as a hyper-parameter using criteria like AIC, BIC, Stein-Unbiased Risk Estimate (SURE)-based approaches or the use techniques presented in \citep{birge2001gaussian}. However, such model selection based schemes often have statistical or computational overheads that make them impractical. The most relevant is the effective degree of freedom (dof) approach (See Eq.(8) and Eq.(9) in \citep{lassodf2}). It requires solving fused lasso with many $\lambda$ (computational overhead). The estimate of dof is unstable in some regimes (statistical overhead). Generally, these methods may work well in practice but often do not come with theoretical guarantees of adaptive optimality. Moreover, we are not aware of any such model-selection technique that can solve the online version of the problem.



There is also a body of work that focuses on the computation of solving  problem~\eqref{eq:fused_penalty} and their higher-dimensional extensions (see \citep{chambolle1997,barbero2011fast}, and the excellent survey therein). This is complementary to our focus, which is to minimize the error against the (unobserved) ground truth. Computationally,  \citep{nickdp}'s dynamic programming has a worst-case $O(n)$ time-complexity, but only for a fixed $\lambda$. Our algorithm runs in $O(n\log n)$-time while avoids choosing the $\lambda$ parameter all together. 



The closest to us is perhaps \citep{arrows2019} which indeed has motivated this work. They consider an online protocol similar to Fig. \ref{fig:online-protocol} with the adversary constrained to reveal the indices $i_t$ isotonically (i.e $i_t = t$) and propose an adaptive restart scheme based on wavelets. However  such techniques are not useful to compete against a more powerful adversary which can query indices in any arbitrary manner --- for example when the exogenous variables $x \in \cX$ are sampled iid from a distribution and revealed online. Further, their proof critically relies on adaptive minimaxity of wavelets. We aim to build a radically new algorithm that is agnostic to the results from wavelet smoothing literature. 

A strongly adaptive online learner \citep{daniely2015strongly,koolen2016specialist},  incurs low static regret in \emph{any} interval. This is accomplished by maintaining a pool of sleeping experts that are static regret minimizing algorithms which are awake only in some  specific duration. Then an aggregation strategy to hedge over the experts is used to guarantee low regret in any interval. This work was preceded by the notion of weakly adaptive regret in \citep{hazan2007adaptive}. To the best of our knowledge, the efficient reduction of TV-denoising to strongly-adaptive online learning is new to this paper. We defer further discussions on related work to Appendix \ref{app:related}.

\section{Preliminaries}
In this section, we briefly review the elements from online learning literature that are crucial to the development of our algorithm.
\subsection{Geometric Cover} \label{sec:gc}
Geometric Cover (GC) proposed in \citep{daniely2015strongly} is a collection of intervals that belong to $\mathbb{N}$ defined below. In what follows $[a,b]$ denotes the set of natural numbers lie between $a$ and $b$, both inclusive.
\begin{align}
    \cI
    &= \bigcup_{k \in \mathbb{N} \cup \{ 0\}} \cI_k,
\end{align}
where $\forall k \in \mathbb{N} \cup \{ 0\}$, and $\cI_k = \{[i \cdot 2^k, (i+1)\cdot 2^k - 1]: i \in \mathbb{N} \}$. Define $\text{AWAKE}(t) := \{I \in \cI: t \in I \}$.By the construction of Geometric Cover $\cI$, it holds that
\begin{align}
    |\text{AWAKE}(t)| = \floor{\log t} + 1. \label{eq:awake}
\end{align}
Let's denote $\cI|_J := \{I \in \cI : I \subseteq J \}$ for an interval $J \subseteq \mathbb{N}$. The GC has a very nice property recorded in the following Proposition.
\begin{proposition} \label{prop:gc}
\citep{daniely2015strongly}  Let $I = [q,s] \subseteq \mathbb{N}$. Then the interval $I$ can be partitioned into two finite sequences of disjoint consecutive intervals $(I_{-k},\ldots,I_0) \subseteq \cI|_I$ and $(I_1,\ldots,I_p) \subseteq \cI|_I$ such that,
\begin{align*}
    \frac{|I_{-i}|}{|I_{-i+1}|} \le \frac{1}{2}, \forall i \ge 1\quad\quad \text{and}\quad\quad     \frac{|I_{i}|}{|I_{i-1}|} \le \frac{1}{2}, \forall i \ge 2.
\end{align*}

\end{proposition}

\subsection{Sleeping Experts and Specialist Aggregation Algorithm (SAA)}
In the problem of learning from expert advice with outcome space $\cO$ and action space $\cA$, there are $K$ experts who provide a list of actions $a_{t,:} = [a_{t,1},...,a_{t,K}]\in \cA^K$ at time $t=1,...,n$. The learner is supposed to takes an action $a_t\in \cA$ based on the expert advice\footnote{Could be $a_{t,k}$ for some $k\in[K]$ or any other points in $\cA$} before the outcome $o_t\in \cO$ is revealed by an adversary. The player then  incurs a loss given by $\ell(a_t, o_t)$, where $\ell$ is a loss function. 

In the most basic setting, $\cA,\cO$ are discrete sets, $\ell$ can be described by a table, and we assign one constant expert to each $a\in\cA$, then this becomes an online version of Von Neumann's linear matrix game. More generally, $\cA$ can be a convex set, describing parameters of a classifier, $o\in\cO$ could denote a feature-label pair in which case the loss could be a square loss or logistic loss that measures the performance of each classifier.

Our result leverages a variant of the learning from expert advice problem which assumes an arbitrary subset of $K$ experts might be sleeping at time $t$ and the learner needs to compete against an expert only during its awake duration. The learner chooses a distribution $\bs w_t$ over the awake experts and plays a weighted average over the actions of those awake experts. It then incurs a surrogate-loss called ``$\mathrm{MixLoss}$'' which is a measure of how good the distribution $\bs w_t$ is. (See Figure~\ref{fig:protocol} for details.) This setting is different from the classical prediction with experts advice problem in two aspects: 1) The adversary is endowed with more power of selecting an awake expert set in addition to the actual outcome $o_t$ at each round. 2) Instead of the loss $\ell(a_t, o_t)$, the learner is incurred a surrogate loss on the distribution chosen by the learner at time $t$.
\begin{figure}[h!]
	\centering
	\fbox{
		\begin{minipage}{7 cm}
		For $t= 1,\ldots,n$
			\begin{enumerate}
				\item Adversary picks a subset $A_t \subset [K]$ of awake experts.
				\item Learner choose a distribution $\bs w_t$ over $A_t$.
				\item Adversary reveals loss of all \emph{awake} experts,\\ $\bs \ell_t \in (-\infty, \infty]^{|A_t|}$. 
				\item Learner suffers $\mathrm{MixLoss}$:\\ $-\log(\sum_{k \in A_t} w_{t,k}e^{-\ell_{t,k}})$.
			\end{enumerate}
		\end{minipage}
	}
	\caption{\emph{Interaction protocol with sleeping experts. The expert pool size is $K$.}}
\label{fig:protocol}
\end{figure}

\begin{figure}[h!]
	\centering
	\fbox{
		\begin{minipage}{7 cm} 
		Initialize $u_{1,k} = 1/|\cS|$ for all $k$ in an index set $\cS$ used to\\ index the expert pool.\\
		   For $t= 1,\ldots,n$
			\begin{enumerate}
			    \item Adversary reveals $A_t \subseteq \cS$.
                \item Play weighted average action wrt distribution:\\ $w_{t,k} = \frac{u_{t,k} \bs 1\{ k \in A_t\}}{\sum_{j \in A_t}u_{t,j}}$.
                \item Broadcast the weights $w_{t,k}$.
                \item Receive losses $\ell_{t,k}$ for all $k \in A_t$.
                \item Update:
                \begin{itemize}
                    \item $u_{t+1,k} = \frac{u_{t,k}e^{-\ell_{t,k}}}{\sum_{j \in A_t}u_{t,j}e^{-\ell_{t,j}}} \sum_{j \in A_t }u_{t,j}$\\ if $k \in A_t$.
                    \item $u_{t+1,k} = u_{t,k}$ if $k \notin A_t$.
                \end{itemize}
			\end{enumerate}
		\end{minipage}
	}
	\caption{\emph{Specialist Aggregation Algorithm (SAA).}}
\label{fig:saa}
\end{figure}

Consider the  protocol of learning with sleeping experts shown in Fig. \ref{fig:protocol}. Assume an expert pool of size $K$.

\begin{lemma} \label{lem:saa}
\citep{koolen2016specialist} Regret $R_n^{j}$ of SAA (Fig. \ref{fig:saa}) w.r.t. any fixed expert $j \in [K]$ satisfies,
\begin{align*}
    R_n^{j} &:= \sum_{t \in [n]} \bs 1\{j \in A_t\} \left( -\log(\sum_{k \in A_t} w_{t,k}e^{-\ell_{t,k}}) - \ell_{t,j}\right)\\
    &\le \log K,
\end{align*}
where $\bs 1\{ \cdot \}$ is the indicator function, $\ell_{t,k} := \cL(a_{t,k},o_t)$ and $a_{t,k}$ is the action taken by expert $k$ at time $t$.
\end{lemma}
Note that $\ell_{t,j}= \mathrm{MixLoss}(\bs{e}_j)$ where $\mathbf{e}_j$ selects $j$ with probability $1$. The regret measures the performance of the learner against any fixed expert in terms of the $\mathrm{MixLoss}$ in the sub-sequence where she is awake.

\begin{definition}
$\cL(a,x)$ is $\eta$ exp-concave in $a$ for each $x$ if $\sum_{k=1}^{K} w_{k}e^{-\eta \cL(a_k,x)} \le e^{-\eta \cL (\sum_{k=1}^{K} w_k a_k,x)}$, for $w_k \ge 0$ and $\sum_{k=1}^{K}w_k = 1$.
\end{definition}

A $\mathrm{MixLoss}$ regret bound is useful because it implies a regret bound on any exp-concave losses for learners playing the weighted average action $a_t = \sum_{k\in A_t} w_{t,k} a_{t,k}$. To see this, let $\cL'(a,o)$ be $\eta$ exp-concave in its first argument $a \in \cA$. By the definition of exp-concavity it follows that  if SAA is run with losses $\cL(a,o) = \eta \cL'(a,o)$, then,
{\small
\begin{align*}
    \sum_{t \in [n] :  j\in A_t} \left( \eta \cL'\left(\sum_{k \in A_t} w_{t,k} a_{t,k}\;, \;o_t \right) - \eta \cL'(a_{t,j},o_t)\right)
    \le R_n^{j},
\end{align*}
}
where $a_{t,k}$ is the action taken by expert $k$ at time $t$.

We refer to Chapter 3 of \citep{BianchiBook2006} and \citep{koolen2016specialist} for further details on SAA. 

\section{Main Results}
In this section, we present our algorithm and its performance guarantees.
\subsection{Algorithm} \label{sec:algo}
As noted in Section \ref{sec:intro}, our goal is to explore the possibility that a Strongly Adaptive online learner can lead to minimax optimal estimation rate. Consequently the algorithm that we present is a fairly standard Strongly Adaptive online learner that can guarantee logarithmic regret in any interval.

Our algorithm \ALG{} (\textbf{A}ggregation of on\textbf{LI}ne avera\textbf{G}es using \textbf{A} geome\textbf{T}ric c\textbf{O}ve\textbf{R}) defined in Fig.\ref{fig:algo} can be used to tackle both online and batch estimation problems. The policy is based on learning with sleeping experts where expert pool is defined as follows.

\begin{definition}
The expert pool is $\cE = \{\cA_I: I \in \cI|_{[n]}\}$, where $\cI|_{[n]}$ is as defined in Section \ref{sec:gc} and $\cA_I$ is an algorithm that perform online averaging in interval $I$. Let $\cA_I(t)$ denote the prediction of the expert $\cA_I$ at time $t$, if $I \in \text{AWAKE}(t)$.
\end{definition}

Due to relation \eqref{eq:awake}, we have $|\cE| \le n \log n$. Our policy basically performs SAA over $\cE$.

\begin{figure}[h!]
	\centering
	\fbox{

		\begin{minipage}{7.5 cm}
\ALG{}:Inputs - time horizon $n$, learning rate $\eta$
\begin{enumerate}
    \item Initialize SAA weights $u_{1,I} = 1/|\cE|, \forall I \in \cI|_{[n]}$.
    \item For $t$ = $1$ to $n$:
    \begin{enumerate}
        \item Adversary reveals an arbitrary $x_{i_t} \in \cX$.
        \item Let $A_t = \text{AWAKE}(i_t)$. Pass $A_t$ to SAA.
        \item Receive $w_{t,I}$ from SAA for each $I \in A_t$.
        \item Predict $\hat{y}_t = \sum_{I \in A_t} w_{t,I} \cA_{I}(t)$.        
        \item Receive $y_t = f(x_{i_t}) + \epsilon_t$.
        \item Pass losses $\ell_{t,I} = \eta (y_t - \cA_{I}(t))^2$,\\ for each $I \in A_t$ to the SAA.
    \end{enumerate}
\end{enumerate}
		\end{minipage}
	}
	\caption{\emph{The \ALG{} algorithm}}
	\label{fig:algo}
\end{figure}

The precise definition of $\cA_I(t)$ used in our algorithm is
\begin{align}
    \cA_I(t)
    &=
    \begin{cases}
    \frac{\sum_{s=1}^{t-1} y_s \bs 1\{i_s \in I\}}{\sum_{s=1}^{t-1} \bs 1\{i_s \in I\}} & \text{ if } \sum_{s=1}^{t-1} \bs 1\{i_s \in I \} > 0\\
    0 & \text{ otherwise }
    \end{cases}
\end{align}
where $i_s$ is the index of the exogenous variable $x_{i_s}$ in step 2(a) of Fig. \ref{fig:algo}. This particular choice of experts is motivated by the fact that performing online averages lead to logarithmic static regret under quadratic losses. As shown later, this property when combined with the SAA scheme leads to logarithmic regret in \emph{any} interval of $[n]$.

\subsection{Performance Guarantees} \label{sec:perf}
\begin{theorem} \label{thm:main}
Consider the online game in Fig. \ref{fig:online-protocol}. Let $\theta_{t} := f(x_{i_t})$. Under Assumption 1, with probability atleast $1-\delta$, \ALG{} forecasts $\hat y_t$ obtained by setting $\eta = \frac{1}{8\left (B+\sigma \sqrt{\log( 2n / \delta)} \right)^2}$, incurs a cumulative error
\begin{align*}
    \sum_{t=1}^{n} (\hat y_t - \theta_t)^2 = \tilde O (n^{1/3}C_n^{2/3}),
\end{align*}
where $\tilde O(\cdot)$ hides the dependency of constants $B, \sigma$ and poly-logarithmic factors of $n$ and $\delta$.
\end{theorem}
\begin{proof}[Proof Sketch]
We first show that \ALG{} suffers logarithmic regret against any expert in the pool $\cE$ during its awake period. Then we exhibit a particular partition of the underlying TV bounded function such that number of chunks in the partition is $O(n^{1/3}C_n^{2/3})$ (Lemma \ref{lem:bins} in Appendix \ref{app:proofs}). Following this, we cover each chunk with atmost $\log n$ experts and show that each expert in the cover suffers a $\tilde O (1)$ estimation error. The Theorem then follows by summing the estimation error across all chunks of the partition. In summary, the delicate interplay between Strongly Adaptive regret bounds and properties of the partition we exhibit leads to the adaptively minimax optimal estimation rate for \ALG{}. We emphasize that existence of such partitions is a highly non-trivial matter.
\end{proof}
\begin{remark}
We note that under the above setting, \ALG{} is minimax optimal in $n$ and $C_n$, and adaptive to unknown $C_n$.
\end{remark}

\begin{remark}
If the noise level $\sigma$ is unknown, it can be robustly estimated from the wavelet coefficients of the observed data by a Median Absolute Deviation estimator \citep{DJBook}. This is facilitated by the sparsity of wavelet coefficients of BV functions .  
\end{remark}

\begin{remark} \label{rem:noparms}
In the offline problem where we have access to all observations ahead of time, the choice of $\eta = 1/(8 \hat \nu^2)$ where $\hat \nu = \max\{|y_1|,\ldots,|y_n|\}$ results in the same near optimal rate for $R_n$ as in Theorem \ref{thm:main}. This is due to the fact that  $B+\sigma \sqrt{\log( 2n / \delta)}$ is nothing but a high probability bound on each $|y_t|$. Hence we don't require the prior knowledge of $B$ and $\sigma$ for the offline problem.
\end{remark}

\begin{remark}
The authors of \citep{donoho1998minimax} use the error metric given by the L2 function norm in a compact interval $[0,1]$ defined as $\int_{0}^{1} \left(\hat f(x) - f(x) \right)^2 dx$ in an offline setting, where $\hat f(x)$ is the estimated function. A common observation model for non-parametric regression considers $x_{i_t} = t/n$ \citep{tibshirani2014adaptive}. When $x_{i_t} = t/n$, \ALG{} guarantees that the empirical norm $\frac{1}{n} \sum_{t=1}^{n} \left (\hat y_t - f(t/n) \right)^2$ decays at the rate of $\tilde O \left ( n^{-2/3} C_n^{2/3}\right)$. For the TV class, it can be shown that the empirical norm and the function norm are close enough such that the estimation rates do not change (see Section 15.5 of \citep{DJBook}).
\end{remark}

\begin{remark}
Note that conditioned on the past observations, the prediction of \ALG{} is deterministic in each round. So in the online setting, we can compete with an adversary who chooses the underlying ground truth in an adaptive manner based on the learner's past moves. With such an adaptive adversary, it becomes important to reveal the set of covariates $\cX$ ahead of time. Otherwise there exists a strategy for the adversary to choose the covariates $x_{i_t}$ that can enforce a linear growth in the cumulative squared error. We refer the readers to \citep{kotlowski2016} for more details about such adversarial strategy.
\end{remark}

\begin{proposition}
The overall run-time of \ALG{} is $O(n \log n)$.
\end{proposition}
\begin{proof}
On each round $|\text{AWAKE}(t)|$ is $O(\log n)$ by \eqref{eq:awake}. So we only need to aggregate and update the weights of $O(\log n)$ experts per round which can be done in $O(\log n)$ time.
\end{proof}

\section{Extensions} \label{sec:ext}
Motivated from a practical perspective, we discus two direct extensions to \ALG{} below. These extensions highlight the versatility of \ALG{} in adapting to each application.

\textbf{Hedged \ALG{}.} In our theoretical results, we found that choosing learning rate $\eta$ conservatively according to Theorem~\ref{thm:main} or Remark~\ref{rem:noparms} ensures the minimax rates. In practice, however, one could use larger learning rates to adapt to the structure of every input sequence.  

We propose to use a hedged \ALG{} scheme that aggregates the predictions of \ALG{} instantiated with different learning rates. In particular, we run different instances of \ALG{} in parallel where an instance corresponds to a learning rate in the exponential grid $[\eta, 2\eta,\ldots,\max\{\eta,\log_2 n \}]$ which has a size of $O\left(\log \left((B^2 + \sigma^2) \log n \right) \right)$. Here $\eta$ is chosen as in Theorem \ref{thm:main} or Remark \ref{rem:noparms}. Then we aggregate each of these instances by the Exponential Weighted Averages (EWA) algorithm \citep{BianchiBook2006}. The learning rate of this outer EWA layer is set according to the theoretical value.
By exp-concavity of squared error losses, this strategy helps to match the performance of the best \ALG{} instance. Since the theoretical choice of learning rate is included in the exponential grid, the strategy can also guarantee optimal minimax rate. 
We emphasize that Hedged \ALG{} is adaptive to $C_n$ and requires no hyper-parameter tuning.


\textbf{\ALG{} with polynomial regression experts.} This extension is motivated by the problem of identifying trends in time series. Though in Section \ref{sec:algo} we use online averaging as experts, in practice one can consider using other algorithms. For example, if the trends in a time series are piecewise-linear, then experts based on online averaging can lead to poor practical performance because the TV budget $C_n$ of piecewise linear signals can be very large. To alleviate this, in this extension, we propose to use Online Polynomial Regression as experts where a polynomial of a fixed degree $d$ is fitted to the data with time points as its exogenous variables. This is similar to the idea adopted in \citep{higherTV} where they construct a policy that performs restarted online polynomial regression where the restart schedule is adaptively chosen via wavelet based methods. They show that such a scheme can guarantee estimation rates that grow with (a scaled) L1 norm of higher order differences of the underlying trend which can be much smaller than its TV budget $C_n$. This extension can be viewed as a variant to the scheme in \citep{higherTV} where the ``hard'' restarts are replaced by ``soft restarts'' via maintaining distributions over the sleeping experts. 

\section{Experimental Results} \label{sec:exp}

\begin{figure}[htp] 
  \centering
  \stackunder{\hspace*{-0.5cm}\includegraphics[width=0.50\textwidth]{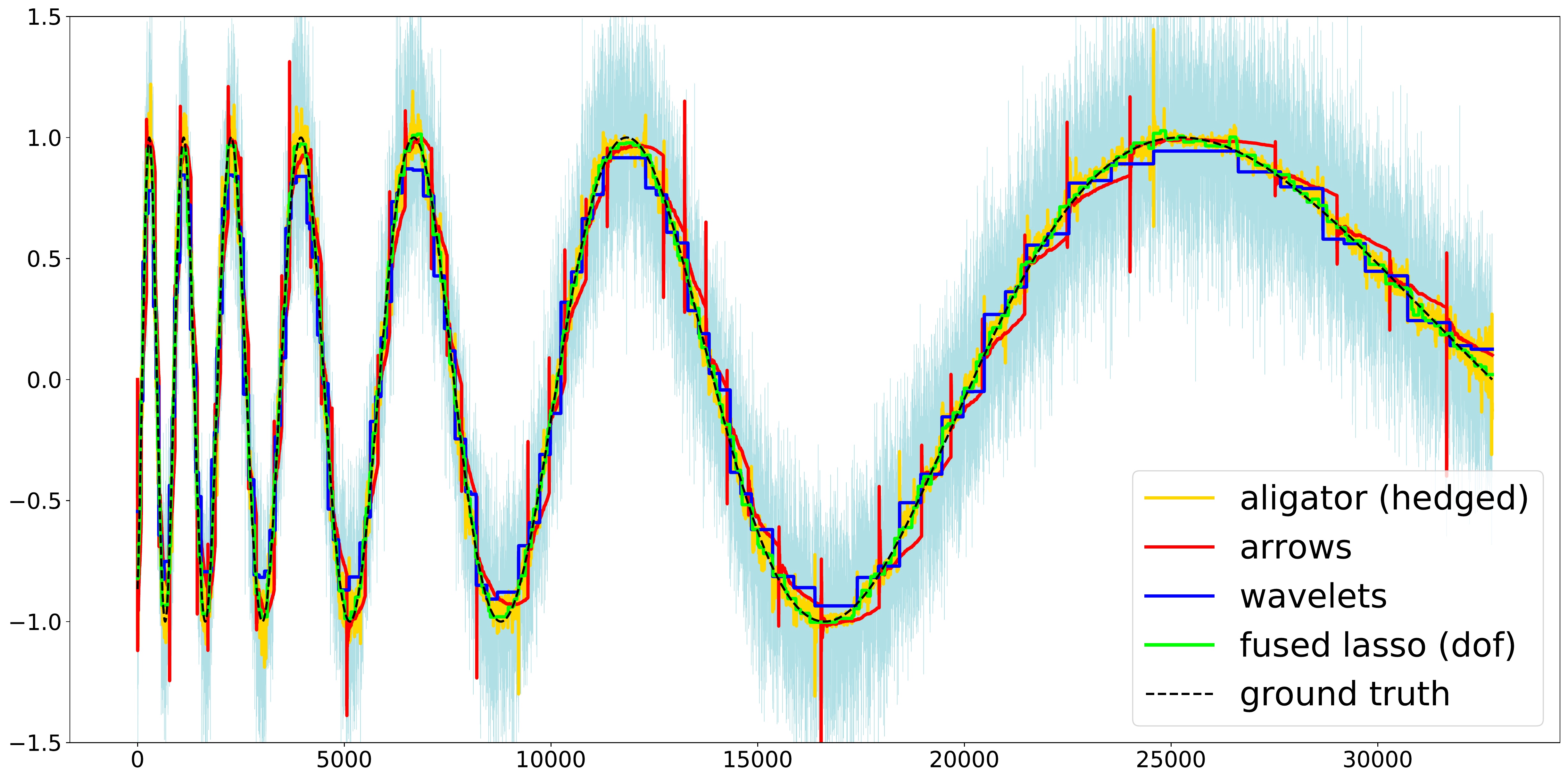}}{}
  \caption{\emph{Fitted signals for Doppler function with noise level $\sigma$= 0.25}}
  \label{fig:dopplerfig}
\end{figure}

\begin{figure}[htp] \label{fig:experiment}  
  \centering
  \stackunder{\hspace*{-0.5cm}\includegraphics[width=0.50\textwidth]{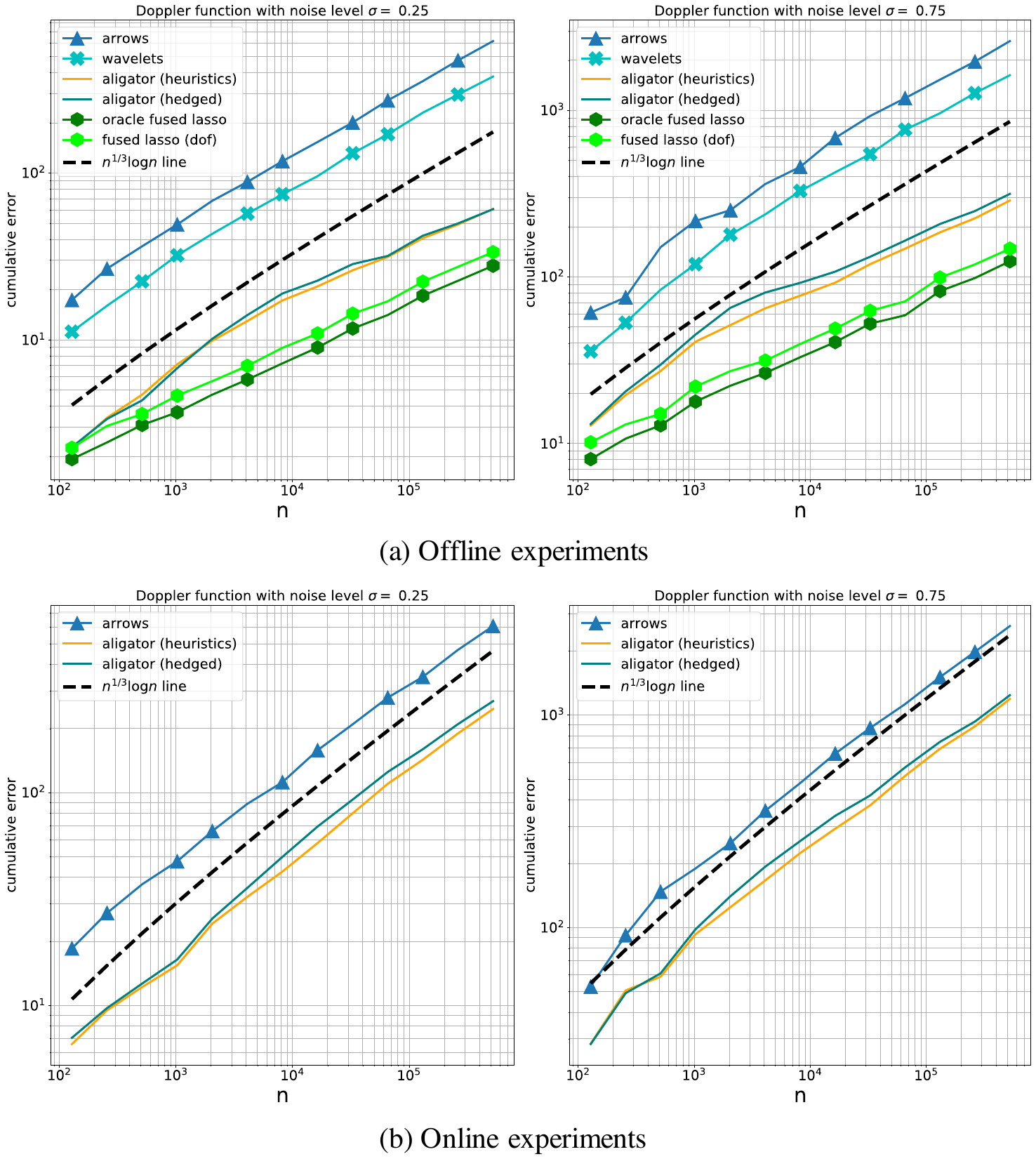}}{}
  \caption{\emph{Cumulative squared error rate of various algorithms on offline setting and online setting. \ALG{} achieves the optimal $\tilde O(n^{1/3})$ rate while performing better than wavelet based methods. In particular, in the offline setting, it achieves a performance closer to that of dof based fused lasso while only incurring a cheap $\tilde O(n)$ run-time overhead.}}
  \label{fig:experiment}
\end{figure}
\begin{figure}[htp]   
  \centering
  \stackunder{\hspace*{-0.5cm}\includegraphics[width=0.50\textwidth,height=0.5\textheight,keepaspectratio]{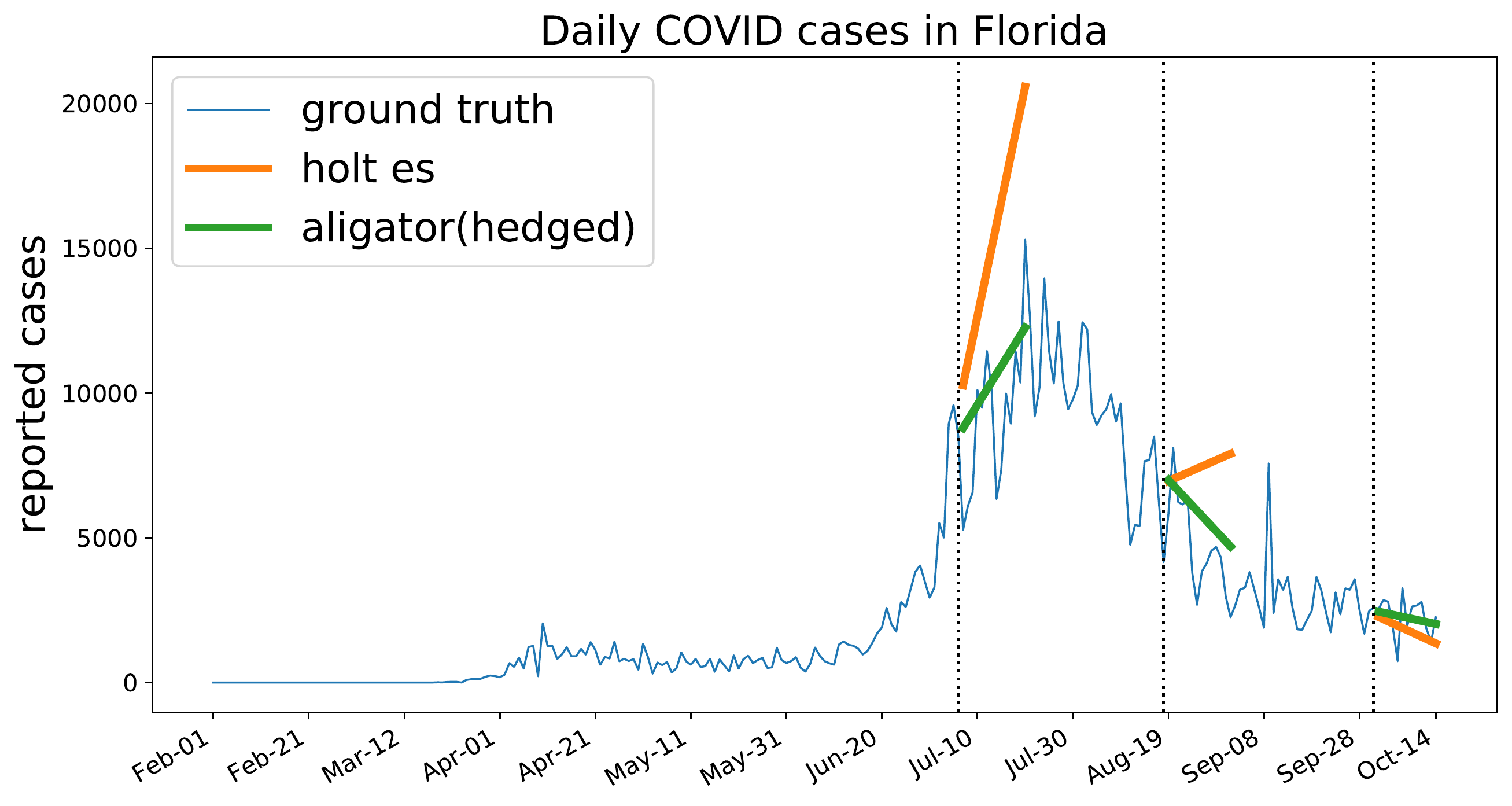}}{}
  \caption{\emph{A demo on forecasting COVID cases based on real world data. We display the two weeks forecasts of hedged \ALG{} and Holt ES, starting from the time points identified by the dotted lines. Both the algorithms are trained on a 2 month data prior to each dotted line. We see that hedged \ALG{}  detects changes in trends more quickly than Holt ES. Further, hedged \ALG{} attains a 20\% reduction in the average RMSE from that of Holt ES (see Section \ref{sec:exp}).}}
  \label{fig:florida}
\end{figure}

For empirical evaluation, we consider online and offline vesrions of the problems separately.

\textbf{Description of policies. } We begin by a description of each algorithm whose error curve is plotted in the figures.

\emph{\ALG{} (hedged):} This is the extension described in Section \ref{sec:ext}

\emph{\ALG{} (heuristics):} For this hueristics strategy, we divide the loss of each expert by $2(\sigma^2 + \sigma^2/m)$ where $m$ is the number of samples whose running average is compued by the expert. This loss is proportional to the notion of (squared) z-score used in hypothesis testing. Intuitively, lower (squared) z-score corresponds to better experts. The multiplier 2 in the previous expression is found to provide good performnace across all signals we consider.

\emph{arrows:} This is the the policy presented in \citep{arrows2019}, which runs online averaging with an adaptive restarting rule based on wavelet denoising results.

\emph{wavelets:} This is the universal soft thresholding estimator from \citep{donoho1998minimax} based on Haar wavelets which is known to be minimax optimal for estimating BV functions.

\emph{oracle fused lasso:} This estimator is obtained by solving \eqref{eq:fused_penalty} whose hyper-parameter is tuned by assuming access to an oracle that can compute the mean squared error wrt  actual ground truth. The exact ranges used in the hyper-parameter grid search is described in Appendix \ref{app:exp}. Note that the oracle fused lasso estimator is purely hypothetical due to absence of such oracles described before in reality and is ultimately impractical. It is used here to facilitate meaningful comparisons.

\emph{fused lasso (dof):} In this experiment, we maintain a list of $\lambda$ for the fused lasso problem (Eq. \eqref{eq:fused_penalty}). Then we compute the Stein's Unbiased Risk estimator for the expected squared error incurred by each $\lambda$ by estimating its degree of freedom (dof) \citep{lassodf2} and select the $\lambda$ with minimum estimated error.

\textbf{Experiments on synthetic data.} For the ground truth signal, we use the Doppler function of \citep{donoho1994ideal} whose waveform is depicted in Fig. \ref{fig:dopplerfig}. The observed data are generated by adding iid noise to the ground truth. For offline setting, we have access to all observations ahead of time. So we run Arrows and both versions of \ALG{} two times on the same data, once in isotonic order (i.e $i_t = t$ in Fig. \ref{fig:online-protocol}) and other in reverse isotonic order and average the predictions to get estimates of the ground truth. For online setting such a forward-backward averaging is not performed.  This process of generating the noisy data and computing estimates are repeated for 5 trials and the average cumulative error is plotted. As we can see from Fig.\ref{fig:experiment} (a), \ALG{} versions attains the $\tilde O(n^{1/3})$ rate and incurs much lower error than wavelet smoothing. Further, performance of hedged and heuristics versions of \ALG{} is in the vicinity to that of the hypothetical fusedlasso estimator while the policies arrows and wavelets violate this property by a large margin. Even though the dof based fused lasso comes very close to the oracle counterpart, we emphasize that this strategy is not known to provide theoretical guarantees for its rate and requires heavy computational bottleneck since it requires to solve the fused lasso (Eq. \ref{eq:fused_penalty}) for many different values of $\lambda$.

For the online version of the problem, we consider the policy Arrows as the benchmark. This policy has been established to be minimax optimal for online forecasting of TV bounded sequences in \citep{arrows2019}. We see from Fig.\ref{fig:experiment} (b) that all the policies attains an $\tilde O(n^{1/3})$ rate while \ALG{} variants enjoy lower cumulative errors.

\textbf{Experiments on real data.} Next we consider the task of forecasting COVID cases using the extension of Aligator with polynomial regression experts as in Section \ref{sec:ext}.  The data are obtained from the CDC website (\cite{cdc}).  

We address a very relevant problem as follows: Given access to the historical data, forecast the evolution of COVID cases for the next 2 weeks. We compare the performance of hedged \ALG{} and Holt Exponential Smoothing (Holt ES), on this problem, where the later is a common algorithm used in Time Series forecasting to detect underlying trends. For \ALG{}, we use Online Linear Regression as experts where a polynomial of degree one is fitted to the data with time points as its exogenous variables. For each time point $t$ in [Apr 20, Sep 27], we train both hedged \ALG{} and Holt ES on a training window of past 2 months. Then we calculate a 2 week forecast for both algorithms. For \ALG{} this is achieved by linearly extrapolating the predictions of experts awake at time $t$ and aggregating them.  Following this, we compute the Root Mean Squared Error (RMSE) in the interval $[t,t+14)$ for both algorithms. These RMSE are then averaged across all $t$ in [Apr 20, Sep 27].

We choose data from the state of Florida, USA, as an illustrative example.
We obtained an average RMSE of 1330.12 for hedged \ALG{} and 1671.77 for Holt ES. Thus hedged \ALG{} attains a 20\% reduction in forecast error from that of Holt ES. A qualitative comparison of the forecasts is illustrated in Fig. \ref{fig:florida}. As we can see, the time series is non-stationary and has a varying degree of smoothness. \ALG{} is able to adapt to the local changes quickly, while Holt ES fails to do so despite having a more sophisticated training phase. Similar experimental results for some of the other states are reported in Appendix \ref{app:exp}.

The training step of hedged \ALG{} involves learning the weights of all experts by an online interaction protocol as shown in Fig. \ref{fig:online-protocol} with $i_t = t$. It is remarkable that \emph{no} hyper-parameter tuning is required by \ALG{} for its training phase. The slowest learning rate to be used in the grid for hedged \ALG{} is computed as follows. First we calculate the maximum loss incurred by each expert for a one step ahead forecast in its awake duration. Then we take the maximum of this quantity across all experts in the pool. Let this quantity be $\beta$. The slowest learning rate in the grid is then set as $1/(2\beta)$. The learning rate of the outer layer of EWA is also set the same. This is justifiable because the quantity $4\left (B+\sigma \sqrt{\log( 2n / \delta)} \right)$ in the denominator of the learning rate in Theorem \ref{thm:main} is a high probability bound on the loss incurred by any expert for a one step ahead forecast.

We defer further experimental results to Appendix \ref{app:exp}.

\textbf{An important caveat for practitioners.} Though \ALG{} is able to detect non-stationary trends in the COVID data efficiently, we do \emph{not} advocate using \ALG{} \emph{as is} for pandemic forecasting, which is a substantially more complex problem that requires input from domain experts. 

However, \ALG{} could have a role in this problem, and other online forecasting tasks.  Estimating (and removing) trend is an important first step in many time series methods (e.g., Box-Jenkins method).  Most trend estimation methods only apply to offline problems (e.g., Hodrick-Prescott filter or L1 Trend Filter) \citep{l1tf}, while Holt ES is a common method used for online trend estimation. For instance, Holt ES is being used as a subroutine for trend estimation in a state-of-the-art forecasting method \citep{jin2020inter} for COVID cases that CDC is currently using. We expect that using \ALG{} instead in such models that use Holt ES will lead to more accurate forecasting, but that is beyond the scope of this paper.


\section{Concluding Discussion}
In this work, we presented a novel reduction from estimating BV functions to Strongly Adaptive online learning. The reduction gives rise to a new algorithm \ALG{} that attains the near minimax optimal rate of $\tilde O(n^{1/3}C_n^{2/3})$ in $O(n \log n)$ run-time. The results form a parallel to wavelet smoothing in terms of optimal adaptivity to unknown variational budget $C_n$. However, our algorithm is more versatile than wavelets in terms of its configurability and practical performance. Further, for offline estimation, \ALG{} variants achieves a performance closer (than wavelets) to an oracle fused lasso while incurring only an $\tilde O(n)$ run-time with no hyper parameter tuning. This is in contrast to degree of freedom based approaches of tuning the fused lasso hyper parameter that requires significantly more computational overhead and is not known to provide guarantees on its rate.


\section*{Acknowledgment}
The research is partially supported by NSF Award \#2029626 and a start-up grant from UCSB CS department.   XZ was supported by UCSB Chancellor's Fellowship.

\bibliography{tf,yx}

\begin{thebibliography}{52}
\providecommand{\natexlab}[1]{#1}
\providecommand{\url}[1]{\texttt{#1}}
\expandafter\ifx\csname urlstyle\endcsname\relax
  \providecommand{\doi}[1]{doi: #1}\else
  \providecommand{\doi}{doi: \begingroup \urlstyle{rm}\Url}\fi

\bibitem[cdc()]{cdc}
Centers for disease control and prevention (cdc).
\newblock \url{https://www.cdc.gov/}.

\bibitem[Adamskiy et~al.(2016)Adamskiy, Koolen, Chernov, and
  Vovk]{koolen2016specialist}
Dmitry Adamskiy, Wouter~M. Koolen, Alexey Chernov, and Vladimir Vovk.
\newblock A closer look at adaptive regret.
\newblock \emph{Journal of Machine Learning Research}, 2016.

\bibitem[Akiyama et~al.(2019)Akiyama, Bouman, and Woody]{blackhole}
Kazunori Akiyama, Katherine Bouman, and David Woody.
\newblock First m87 event horizon telescope results. i. the shadow of the
  supermassive black hole.
\newblock \emph{Astrophysical Journal Letters}, 2019.

\bibitem[Baby and Wang(2019)]{arrows2019}
Dheeraj Baby and Yu-Xiang Wang.
\newblock Online forecasting of total-variation-bounded sequences.
\newblock In \emph{Neural Information Processing Systems (NeurIPS)}, 2019.

\bibitem[Baby and Wang(2020)]{higherTV}
Dheeraj Baby and Yu-Xiang Wang.
\newblock Adaptive online estimation of piecewise polynomial trends.
\newblock \emph{To appear in Advances in Neural Information Processing Systems
  (NeurIPS)}, 2020.

\bibitem[Barbero and Sra(2011)]{barbero2011fast}
Alvaro Barbero and Suvrit Sra.
\newblock Fast {Newton-type} methods for total variation regularization.
\newblock In \emph{International Conference on Machine Learning (ICML-11)},
  volume~28, pages 313--320, 2011.

\bibitem[Besbes et~al.(2015)Besbes, Gur, and Zeevi]{besbes2015non}
Omar Besbes, Yonatan Gur, and Assaf Zeevi.
\newblock Non-stationary stochastic optimization.
\newblock \emph{Operations research}, 63\penalty0 (5):\penalty0 1227--1244,
  2015.

\bibitem[Beygelzimer et~al.(2011)Beygelzimer, Langford, Li, Reyzin, and
  Schapire]{beygel2011contextual}
Alina Beygelzimer, John Langford, Lihong Li, Lev Reyzin, and Robert Schapire.
\newblock Contextual bandit algorithms with supervised learning guarantees.
\newblock In \emph{Proceedings of the Fourteenth International Conference on
  Artificial Intelligence and Statistics}, 2011.

\bibitem[Birge and Massart(2001)]{birge2001gaussian}
Lucien Birge and Pascal Massart.
\newblock Gaussian model selection.
\newblock \emph{Journal of the European Mathematical Society}, 3\penalty0
  (3):\penalty0 203--268, 2001.

\bibitem[Cesa-Bianchi and Lugosi(2006)]{BianchiBook2006}
Nicolo Cesa-Bianchi and Gabor Lugosi.
\newblock \emph{Prediction, Learning, and Games}.
\newblock Cambridge University Press, New York, NY, USA, 2006.
\newblock ISBN 0521841089.

\bibitem[Chambolle and Lions(1997)]{chambolle1997}
Antonin Chambolle and Pierre-Louis Lions.
\newblock Image recovery via total variation minimization and related problems.
\newblock \emph{Numerische Mathematik}, 76\penalty0 (2):\penalty0 167--188,
  1997.

\bibitem[Chambolle et~al.(2010)Chambolle, Caselles, Cremers, Novaga, and
  Pock]{chambolle2010introduction}
Antonin Chambolle, Vicent Caselles, Daniel Cremers, Matteo Novaga, and Thomas
  Pock.
\newblock An introduction to total variation for image analysis.
\newblock \emph{Theoretical foundations and numerical methods for sparse
  recovery}, 9\penalty0 (263-340):\penalty0 227, 2010.

\bibitem[Chen et~al.(2018{\natexlab{a}})Chen, Goel, and
  Wierman]{chen2018smoothed}
Niangjun Chen, Gautam Goel, and Adam Wierman.
\newblock Smoothed online convex optimization in high dimensions via online
  balanced descent.
\newblock In \emph{Conference on Learning Theory (COLT-18)},
  2018{\natexlab{a}}.

\bibitem[Chen et~al.(2018{\natexlab{b}})Chen, Wang, and Wang]{chen2018non}
Xi~Chen, Yining Wang, and Yu-Xiang Wang.
\newblock Non-stationary stochastic optimization under lp, q-variation
  measures.
\newblock 2018{\natexlab{b}}.

\bibitem[Daniely et~al.(2015)Daniely, Gonen, and
  Shalev-Shwartz]{daniely2015strongly}
Amit Daniely, Alon Gonen, and Shai Shalev-Shwartz.
\newblock Strongly adaptive online learning.
\newblock In \emph{International Conference on Machine Learning}, pages
  1405--1411, 2015.

\bibitem[Donoho and Johnstone(1994{\natexlab{a}})]{donoho1994ideal}
David Donoho and Iain Johnstone.
\newblock Ideal spatial adaptation by wavelet shrinkage.
\newblock \emph{Biometrika}, 81\penalty0 (3):\penalty0 425--455,
  1994{\natexlab{a}}.

\bibitem[Donoho and Johnstone(1994{\natexlab{b}})]{donoho1994minimax}
David Donoho and Iain Johnstone.
\newblock Minimax risk over $\ell_p$-balls for $\ell_q$-error.
\newblock \emph{Probability Theory and Related Fields}, 99\penalty0
  (2):\penalty0 277--303, 1994{\natexlab{b}}.

\bibitem[Donoho et~al.(1990)Donoho, Liu, and MacGibbon]{donoho1990minimax}
David Donoho, Richard Liu, and Brenda MacGibbon.
\newblock Minimax risk over hyperrectangles, and implications.
\newblock \emph{Annals of Statistics}, 18\penalty0 (3):\penalty0 1416--1437,
  1990.

\bibitem[Donoho et~al.(1998)Donoho, Johnstone, et~al.]{donoho1998minimax}
David~L Donoho, Iain~M Johnstone, et~al.
\newblock Minimax estimation via wavelet shrinkage.
\newblock \emph{The annals of Statistics}, 26\penalty0 (3):\penalty0 879--921,
  1998.

\bibitem[Guntuboyina et~al.(2017)Guntuboyina, Lieu, Chatterjee, and
  Sen]{guntuboyina2018constrainedTF}
Adityanand Guntuboyina, Donovan Lieu, Sabyasachi Chatterjee, and Bodhisattva
  Sen.
\newblock Adaptive risk bounds in univariate total variation denoising and
  trend filtering.
\newblock 2017.

\bibitem[Hall and Willett(2013)]{hall2013dynamical}
Eric Hall and Rebecca Willett.
\newblock Dynamical models and tracking regret in online convex programming.
\newblock In \emph{International Conference on Machine Learning (ICML-13)},
  pages 579--587, 2013.

\bibitem[Haussler et~al.(1998)Haussler, Kivinen, and
  Warmuth]{Haussler1998mixloss}
D.~Haussler, J.~Kivinen, and M.~K. Warmuth.
\newblock Sequential prediction of individual sequences under general loss
  functions.
\newblock \emph{IEEE Trans. Inf. Theor.}, 1998.

\bibitem[Hazan and Seshadhri(2007)]{hazan2007adaptive}
Elad Hazan and Comandur Seshadhri.
\newblock Adaptive algorithms for online decision problems.
\newblock In \emph{Electronic colloquium on computational complexity (ECCC)},
  volume~14, 2007.

\bibitem[Hoefling(2010)]{hoefling2010path}
Holger Hoefling.
\newblock A path algorithm for the fused lasso signal approximator.
\newblock \emph{Journal of Computational and Graphical Statistics}, 19\penalty0
  (4):\penalty0 984--1006, 2010.

\bibitem[Hutter and Rigollet(2016)]{hutter2016optimal}
Jan-Christian Hutter and Philippe Rigollet.
\newblock Optimal rates for total variation denoising.
\newblock In \emph{Conference on Learning Theory (COLT-16)}, 2016.

\bibitem[Jadbabaie et~al.(2015)Jadbabaie, Rakhlin, Shahrampour, and
  Sridharan]{jadbabaie2015online}
Ali Jadbabaie, Alexander Rakhlin, Shahin Shahrampour, and Karthik Sridharan.
\newblock Online optimization: Competing with dynamic comparators.
\newblock In \emph{Artificial Intelligence and Statistics}, pages 398--406,
  2015.

\bibitem[Jin et~al.(2021)Jin, Wang, and Yan]{jin2020inter}
Xiaoyong Jin, Yu-Xiang Wang, and Xifeng Yan.
\newblock Inter-series attention model for covid-19 forecasting.
\newblock \emph{SIAM International Conference on Data Mining (to appear)},
  2021.

\bibitem[Johnson(2013)]{nickdp}
Nicholas Johnson.
\newblock A dynamic programming algorithm for the fused lasso and
  {$L_0$}-segmentation.
\newblock \emph{Journal of Computational and Graphical Statistics}, 22\penalty0
  (2):\penalty0 246--260, 2013.

\bibitem[Johnstone(2017)]{DJBook}
Iain~M. Johnstone.
\newblock \emph{Gaussian estimation: Sequence and wavelet models}.
\newblock 2017.

\bibitem[Kim et~al.(2009)Kim, Koh, Boyd, and Gorinevsky]{l1tf}
Seung-Jean Kim, Kwangmoo Koh, Stephen Boyd, and Dimitry Gorinevsky.
\newblock $\ell_1$ trend filtering.
\newblock \emph{SIAM Review}, 51\penalty0 (2):\penalty0 339--360, 2009.

\bibitem[Kotłowski et~al.(2016)Kotłowski, Koolen, and Malek]{kotlowski2016}
Wojciech Kotłowski, Wouter~M. Koolen, and Alan Malek.
\newblock Online isotonic regression.
\newblock In \emph{Annual Conference on Learning Theory (COLT-16)}, volume~49,
  pages 1165--1189. PMLR, 2016.

\bibitem[Lee et~al.(2019)Lee, Gommers, Waselewski, Wohlfahrt, and
  Leary]{pywavelets}
Gregory Lee, Ralf Gommers, Filip Waselewski, Kai Wohlfahrt, and Aaron~O Leary.
\newblock Pywavelets: A python package for wavelet analysis.
\newblock \emph{Journal of Open Source Software}, 2019.
\newblock URL \url{https://doi.org/10.21105/joss.01237}.

\bibitem[Li et~al.(2018)Li, Mark, Raskutti, and Willett]{li2018graph}
Yuan Li, Benjamin Mark, Garvesh Raskutti, and Rebecca Willett.
\newblock Graph-based regularization for regression problems with
  highly-correlated designs.
\newblock In \emph{2018 IEEE Global Conference on Signal and Information
  Processing (GlobalSIP)}, pages 740--742. IEEE, 2018.

\bibitem[Mammen(1991)]{mammen1991}
Enno Mammen.
\newblock Nonparametric regression under qualitative smoothness assumptions.
\newblock \emph{Annals of Statistics}, 19\penalty0 (2):\penalty0 741---759,
  1991.

\bibitem[Mammen and van~de Geer(1997)]{locadapt}
Enno Mammen and Sara van~de Geer.
\newblock Locally apadtive regression splines.
\newblock \emph{Annals of Statistics}, 25\penalty0 (1):\penalty0 387--413,
  1997.

\bibitem[Padilla et~al.(2017)Padilla, Sharpnack, and Scott]{padilla2017dfs}
Oscar Hernan~Madrid Padilla, James Sharpnack, and James~G Scott.
\newblock The dfs fused lasso: Linear-time denoising over general graphs.
\newblock \emph{The Journal of Machine Learning Research}, 18\penalty0
  (1):\penalty0 6410--6445, 2017.

\bibitem[Rakhlin and Sridharan(2014)]{rakhlin2014online}
Alexander Rakhlin and Karthik Sridharan.
\newblock Online non-parametric regression.
\newblock In \emph{Conference on Learning Theory}, pages 1232--1264, 2014.

\bibitem[Rudin et~al.(1992)Rudin, Osher, and Faterni]{tv}
Leonid Rudin, Stanley Osher, and Emad Faterni.
\newblock Nonlinear total variation based noise removal algorithms.
\newblock \emph{Physica {D}: Nonlinear Phenomena}, 60:\penalty0 259--268, 1992.

\bibitem[Sadhanala et~al.(2016{\natexlab{a}})Sadhanala, Wang, and
  Tibshirani]{sadhanala2016total}
Veeranjaneyulu Sadhanala, Yu-Xiang Wang, and Ryan Tibshirani.
\newblock Total variation classes beyond 1d: Minimax rates, and the limitations
  of linear smoothers.
\newblock \emph{Advances in Neural Information Processing Systems (NIPS-16)},
  2016{\natexlab{a}}.

\bibitem[Sadhanala et~al.(2016{\natexlab{b}})Sadhanala, Wang, and
  Tibshirani]{sadhanala2016graph}
Veeranjaneyulu Sadhanala, Yu-Xiang Wang, and Ryan~J Tibshirani.
\newblock Graph sparsification approaches for laplacian smoothing.
\newblock In \emph{AISTATS'16}, pages 1250--1259, 2016{\natexlab{b}}.

\bibitem[Sadhanala et~al.(2017)Sadhanala, Wang, Sharpnack, and
  Tibshirani]{sadhanala2017higher}
Veeranjaneyulu Sadhanala, Yu-Xiang Wang, James Sharpnack, and Ryan Tibshirani.
\newblock Higher-order total variation classes on grids: Minimax theory and
  trend filtering methods.
\newblock \emph{Advances in Neural Information Processing Systems (NIPS-17)},
  2017.

\bibitem[Tibshirani et~al.(2005)Tibshirani, Saunders, Rosset, Zhu, and
  Knight]{fuse}
Robert Tibshirani, Michael Saunders, Saharon Rosset, Ji~Zhu, and Keith Knight.
\newblock Sparsity and smoothness via the fused lasso.
\newblock \emph{Journal of the Royal Statistical Society: Series B},
  67\penalty0 (1):\penalty0 91--108, 2005.

\bibitem[Tibshirani(2014)]{tibshirani2014adaptive}
Ryan~J Tibshirani.
\newblock Adaptive piecewise polynomial estimation via trend filtering.
\newblock \emph{Annals of Statistics}, 42\penalty0 (1):\penalty0 285--323,
  2014.

\bibitem[Tibshirani and Taylor(2012)]{lassodf2}
Ryan~J. Tibshirani and Jonathan Taylor.
\newblock Degrees of freedom in lasso problems.
\newblock \emph{Annals of Statistics}, 40\penalty0 (2):\penalty0 1198--1232,
  2012.

\bibitem[Tsybakov(2008)]{tsybakov_book}
Alexandre~B. Tsybakov.
\newblock \emph{Introduction to Nonparametric Estimation}.
\newblock Springer Publishing Company, Incorporated, 1st edition, 2008.

\bibitem[Wang et~al.(2016)Wang, Sharpnack, Smola, and Tibshirani]{graphtf}
Yu-Xiang Wang, James Sharpnack, Alex Smola, and Ryan~J Tibshirani.
\newblock Trend filtering on graphs.
\newblock \emph{Journal of Machine Learning Research}, 17\penalty0
  (105):\penalty0 1--41, 2016.

\bibitem[Yang et~al.(2016)Yang, Zhang, Jin, and Yi]{yang2016tracking}
Tianbao Yang, Lijun Zhang, Rong Jin, and Jinfeng Yi.
\newblock Tracking slowly moving clairvoyant: optimal dynamic regret of online
  learning with true and noisy gradient.
\newblock In \emph{International Conference on Machine Learning (ICML-16)},
  pages 449--457, 2016.

\bibitem[Yuan and Lamperski(2019)]{yuan2019dynamic}
Jianjun Yuan and Andrew Lamperski.
\newblock Trading-off static and dynamic regret in online least-squares and
  beyond.
\newblock 2019.

\bibitem[{Zhang} et~al.(2017){Zhang}, {Zuo}, {Chen}, {Meng}, and
  {Zhang}]{dldenoiser}
K.~{Zhang}, W.~{Zuo}, Y.~{Chen}, D.~{Meng}, and L.~{Zhang}.
\newblock Beyond a gaussian denoiser: Residual learning of deep cnn for image
  denoising.
\newblock \emph{IEEE Transactions on Image Processing}, 2017.

\bibitem[Zhang et~al.(2018{\natexlab{a}})Zhang, Lu, and
  Zhou]{zhang2018adaptive}
Lijun Zhang, Shiyin Lu, and Zhi-Hua Zhou.
\newblock Adaptive online learning in dynamic environments.
\newblock In \emph{Advances in Neural Information Processing Systems
  (NeurIPS-18)}, pages 1323--1333, 2018{\natexlab{a}}.

\bibitem[Zhang et~al.(2018{\natexlab{b}})Zhang, Yang, Zhou,
  et~al.]{zhang2018dynamic}
Lijun Zhang, Tianbao Yang, Zhi-Hua Zhou, et~al.
\newblock Dynamic regret of strongly adaptive methods.
\newblock In \emph{International Conference on Machine Learning (ICML-18)},
  pages 5877--5886, 2018{\natexlab{b}}.

\bibitem[Zinkevich(2003)]{zinkevich2003online}
Martin Zinkevich.
\newblock Online convex programming and generalized infinitesimal gradient
  ascent.
\newblock In \emph{International Conference on Machine Learning (ICML-03)},
  pages 928--936, 2003.

\end{thebibliography}
\bibliographystyle{plainnat}

\newpage
\onecolumn

\appendix

\section{More on Related Work}\label{app:related}

For any forecasting strategy whose output $\hat{y}_t$ at time $t$ depends only on past observations, we have $\mathbb{E}[(\hat y_t - y_t)^2] - \mathbb{E}[(f(x_{i_t}) - y_t)^2] = \mathbb{E}[(\hat y_t - f(x_{i_t}))^2]$. Hence any algorithm that minimizes the dynamic regret against the sequence $f(x_{i_1}),\ldots,f(x_{i_n})$ with $\ell_t(x) = (x - y_t)^2$ being the loss at time $t$, can be potentially applied to solve our problem. However as noted in \citep{arrows2019} a wide array of techniques such as \citep{zinkevich2003online,hall2013dynamical,besbes2015non,chen2018non,jadbabaie2015online,yang2016tracking,zhang2018adaptive,zhang2018dynamic,chen2018smoothed, yuan2019dynamic} are unable to achieve the optimal rate. However, we note that many of these algorithms support general convex/strongly-convex losses. The existence of a strategy with $\tilde O(n^{1/3}C_n^{2/3})$ rate for $R_n$, even in the more general (in comparison to offline problem) online setting considered in Fig. \ref{fig:online-protocol} is implied by the results of \citep{rakhlin2014online} on online non-parametric regression with Besov spaces via a non-constructive argument. \citep{kotlowski2016} studies the problem of forecasting isotonic sequences. However, the techniques are not extensible to forecasting the much richer family of TV bounded sequences.

We acknowledge that univariate TV-denoising is a simple and classical problem setting, and there had been a number of studies on TV-denoising in multiple dimensions and on graphs, and to higher order TV functional, while establishing the optimal rates in those settings \citep{tibshirani2014adaptive,graphtf,hutter2016optimal,sadhanala2016total,sadhanala2017higher,li2018graph}. The problem of adaptivity in $C_n$ is generally open for those settings, except for highly special cases where the optimal tuning parameter happens to be independent to $C_n$ (see e.g., \citep{hutter2016optimal}). Generalization of the techniques developed in this paper to these settings are possible but beyond the scope of this paper. That said, as \citep{padilla2017dfs} establishes, an adaptive univariate fused lasso is already able to handle signal processing tasks on graphs with great generality by simply taking the depth-first-search order as a chain. 

Using a specialist aggregation scheme to incur low adaptive regret was  explored in \citep{koolen2016specialist}. However, the experts they use are same as that of \citep{hazan2007adaptive}. Due to this, their techniques are not directly applicable in our setting where the exogenous variables are queried in an arbitrary manner.

There are image denoising algorithms based on deep neural networks such as \citep{dldenoiser}. However, this body of work is complementary to our focus on establishing the connection between denoising and strongly adaptive online learning.

\section{Proofs of Technical Results} \label{app:proofs}

For the sake of clarity, we  present a sequence of lemmas and sketch how to chain them to reach the main result in Section \ref{part1}. This is followed by proof of all lemmas in Section \ref{part2} and finally the proof of Theorem \ref{thm:main} in Section \ref{part3}. 

\subsection{Proof strategy for Theorem \ref{thm:main}} \label{part1}

We first show that \ALG{} suffers logarithmic regret against any expert in the pool $\cE$ during its awake period. Then we exhibit a particular partition of the underlying TV bounded function such that number of chunks in the partition is $O(n^{1/3}C_n^{2/3})$. Following this, we cover each chunk with atmost $\log n$ experts and show that each expert in the cover suffers a $\tilde O (1)$ estimation error. The Theorem then follows by summing the estimation error across all chunks.

\textbf{Some notations.} In the analysis thereafter, we will use the following notations. Let $\tilde \sigma =  \sigma \sqrt{2 \log (4n/\delta)}$, $R_\sigma = 16(B+\tilde \sigma)^2$ and $\cT(I) = \{t \in [n]: i_t \in I \}$ for any $I \in \cI|_{[n]}$, where $\cI|_{[n]}$ is defined according to the terminology in Section \ref{sec:gc}. Let $\theta_t := f(x_{i_t})$.

First, we show that \ALG{} is competitive against any expert in the pool $\cE$.
\begin{restatable}{lemma}{lemmstrongadapt} \label{lem:strong-adapt}
For any interval $I \in \cI|_{[n]}$ such that $\cT(I)$ is non-empty, the predictions made by \ALG{} $\hat{y}_t$ satisfy
\begin{align}
    \sum_{t \in \cT(I)} (\hat y_t - \theta_t)^2
    &\le \frac{e-1}{3-e} \sum_{t \in \cT(I)} (\cA_I(t) - \theta_t)^2 + \frac{\log(n \log n) R_\sigma + 2 R_\sigma^2 \log(2n \log n/\delta)}{3-e},
\end{align}
with probability atleast $1-\delta$.
\end{restatable}

\begin{corollary} \label{cor:bins}
Let $\mathbb{S} = \{P_1,\ldots, P_M\}$ be an arbitrary ordered set of consecutive intervals in $[n]$. For each $i \in [n]$ let $\mathcal{U}_i$ be the set containing elements of the GC that covers the interval $P_i$ according to Proposition \ref{prop:gc}. Denote $\lambda :=  \frac{\log(n \log n) R_\sigma + 2 R_\sigma^2 \log(2n \log n/\delta)}{3-e}$. Then \ALG{} forecasts $\hat y_t$ satisfy
\begin{align}
    \sum_{t=1}^n (\hat y_t - \theta_t)^2
    &\le \min_{\mathbb{S}} \sum_{i=1}^M\sum_{I \in \mathcal{U}_i} \bs 1\{|\cT(I)| > 0\} \left( \frac{e-1}{3-e} \sum_{t \in \cT(I)} (\cA_I(t) - \theta_t)^2 + \lambda \right),
\end{align}
with probability atleast $1-\delta$.
\end{corollary}
The minimum across all partitions in the Corollary above hints to the novel ability of \ALG{} to incur potentially very low estimation errors.

Next, we proceed to exhibit a partition of the set of exogenous variables queried by the adversary that will eventually lead to the minimax rate of $\tilde O(n^{1/3}C_n^{2/3})$. The existence of such partitions is a non-trivial matter.

\begin{restatable}{lemma}{lemmbins} \label{lem:bins}
Let $ \mathbb{S} = \{x_{k_1} < \ldots, < x_{k_m}\} \subseteq \cX $ be the exogenous variables queried by the adversary over $n$ rounds where each $k_i \in [n]$. Denote $\theta^{(i)} := f(x_{k_i})$ and $p(i) := \#\{t: x_{i_t} = x_{k_i} \}$ for each $i \in [m]$. Denote $[x_i,x_j] := \{x_{k_i},x_{k_{i+1}},\ldots,x_{k_j}\}$. For any  $[x_i,x_j] \subseteq \mathbb{S}$, define $V(x_i,x_j) = \sum_{k=i}^{j-1} |\theta^{(i)} - \theta^{(i+1)} |$. There exists a partitioning $\mathcal{P} = \{ [x_1,x_{r_1}], [x_{r_1+1}, x_{r_2}], \ldots, [x_{r_{M-1}+1},x_{m}]\}$ of $\mathbb{S}$ that satisfies
\begin{enumerate}
    \item For any $[x_i, x_j] \in \mathcal{P} \setminus \{[x_{r_{M-1}+1},x_m] \}$, $V(x_i,x_j) \le \frac{B}{\sqrt{\sum_{k=i}^{j} p(k)}}$.
    \item $V(x_{r_{M-1}+1},x_{m-1}) \le \frac{B}{\sqrt{\sum_{k=r_{M-1}+1}^{m-1} p(k)}}$. 
    \item Number of partitions $M \le \max\{3n^{1/3}C_n^{2/3}B^{-2/3},1\}$.
\end{enumerate}
\end{restatable}

The next lemma controls the estimation error incurred by an expert during its awake period.

\begin{restatable}{lemma}{lemmexp} \label{lem:expert-concent}
Let $\{\ubar{x}, < \ldots, < \bar{x}\}$ be the exogenous variables queried by the adversary over $n$ rounds in an arbitrary interval $I \in \cI|_{[n]}$. Then with probability atleast $1-\delta$
\begin{align}
    \sum_{t \in \cT(I)} (\theta_t - \cA_I(t))^2
    &\le 2V(\ubar{x},\bar{x})^2 |\cT(I)| +  2\sigma^2 \log (2n^3 \log n/\delta) \log(|\cT(I)|),
\end{align}
where $V(\cdot,\cdot)$ is defined as in Lemma~\ref{lem:bins}.
\end{restatable}

To prove Theorem \ref{thm:main}, our strategy is to apply Corollary  \ref{cor:bins} to the partition in Lemma~\ref{lem:bins}. By the construction of the GC, each chunk in the partition can be covered using atmost $\log n$ intervals. Now consider the estimation error incurred by an expert corresponding to one such interval. Due to statements 1 and 2 in Lemma~\ref{lem:bins} the $V(\ubar{x},\bar{x})^2 |\cT(I)|$ term of error bound in Lemma~\ref{lem:expert-concent} can be shown to $O(1)$. When summed across all intervals that cover a chunk, the total estimation error within a chunk becomes $\tilde O(1)$. Now appealing to statement 3 of Lemma~\ref{lem:bins}, we get a total error of $\tilde O(n^{1/3}C_n^{2/3})$ when the error is summed across all chunks in the partition.

\subsection{Omitted Lemmas and Proofs} \label{part2}
\begin{lemma} \label{lem:ub}
Let $\mathcal{V}$ be the event that for all $t \in [n]$, $|\epsilon_t| \le \sigma \sqrt{2 \log (4n/\delta)}$. Then $\mathbb{P}(\mathcal{V}) \ge 1-\delta/2$.
\end{lemma}
\begin{proof}
By gaussian tail inequality, we have for a fixed $t$ $P(|\epsilon_t| > \sigma \sqrt{2 \log (4n/\delta)} ) \le \delta/2n $. By taking a union bound we get $P(|\epsilon_t| \ge \sigma \sqrt{2 \log (4n/\delta)} ) \le \delta/2$ for all $t \in [n]$.
\end{proof}

\textbf{Some notations.} In the analysis thereafter, we will use the following filtration.
\begin{align}
    \cF_j = \sigma((i_1,y_{i_1}),\ldots,(i_{j-1},y_{i_{j-1}})).
\end{align}
Let's denote $\mathbb{E}_{j}[\cdot] := \mathbb{E}[\cdot| \cF_j]$ and $\Var_j[\cdot] := \Var[\cdot | \cF_j]$. Let $\theta_{j} = f(x_{i_j})$ and $\tilde \sigma =  \sigma \sqrt{2 \log (4n/\delta)}$. Let $R_\sigma = 16(B+\tilde \sigma)^2$ and $\cT(I) = \{t \in [n]: i_t \in I \}$

\begin{lemma} \label{lem:freedman}
(Freedman type inequality, \citep{beygel2011contextual}) For any real valued martingale difference sequence $\{ Z _t\}_{t=1}^T$ with $|Z_t| \le R$ it holds that,
\begin{align}
    \sum_{t=1}^{T} Z_t \le \eta(e-2) \sum_{t=1}^T \Var_t[Z_t] + \frac{R \log(1/\delta)}{\eta},
\end{align}
with probability atleast $1-\delta$ for all $\eta \in [0,1/R]$.
\end{lemma}

\begin{lemma} \label{lem:selfbound}
For any $j \in [n]$, we have
\begin{enumerate}
    \item $\mathbb{E}_{j}[(y_j - \cA_I(j) )^2 - (y_j - \theta_{j})^2 | \mathcal{V}] = \mathbb{E}_{j}[(\cA_I(j) - \theta_j)^2|\mathcal{V}]$.
    \item $\Var_{j}[(y_j - \cA_I(j) )^2 - (y_j - \theta_{j})^2|\mathcal{V}] \le R_\sigma \mathbb{E}_{j}[(\cA_I(j) - \theta_j)^2|\mathcal{V}]$.
\end{enumerate}
\end{lemma}
\begin{proof}
We have,
\begin{align}
    \mathbb{E}_{j}[(y_j - \cA_I(j) )^2 - (y_j - \theta_{j})^2|\mathcal{V}]
    &=_{(a)} \mathbb{E}_{j}[(\cA_I(j) - \theta_j)^2|\mathcal{V}] - 2 \mathbb{E}_{j}[\epsilon_j|\mathcal{V}] \mathbb{E}_{j}[(\cA_I(j) - \theta_j)|\mathcal{V}],\\
    &= \mathbb{E}_{j}[(\cA_I(j) - \theta_j)^2|\mathcal{V}],
\end{align}
where line (a) is due to the independence of $\epsilon_j$ with the past. Since $(\cA_I(j) + \theta_{j} - 2y_j)^2 \le 16(B+\tilde \sigma)^2$ under the event $\mathcal{V}$, it holds that
\begin{align}
    \Var_{j}[(y_j - \cA_I(j) )^2 - (y_j - \theta_{j})^2|\mathcal{V}]
    &\le \mathbb{E}_{j}[(y_j - \cA_I(j) )^2 - (y_j - \theta_{j})^2|\mathcal{V}]^2,\\
    &\le 16(B+\tilde \sigma)^2 \mathbb{E}_{j}[(\cA_I(j) - \theta_j)^2|\mathcal{V}].
\end{align}

\end{proof}

\begin{lemma} \label{lem:prob}
For any interval $I \in \cI$, it holds with probability atleast $1-\delta$ that
\begin{enumerate}
    \item $\sum_{j \in \cT(I)} (y_j - \cA_I(j) )^2 - (y_j - \theta_{j})^2 \le \sum_{j \in \cT(I)} (e-1)(\cA_I(j) - \theta_j)^2 + R_\sigma^2 \log(2n \log n/\delta)$,
    \item $\sum_{j \in \cT(I)} (y_j - \hat{y}_j )^2 - (y_j - \theta_{j})^2 \ge \sum_{j \in \cT(I)} (3-e) (\hat{y}_j - \theta_j)^2 - R_\sigma^2 \log(2n \log n/\delta)$.
\end{enumerate}
\end{lemma}
\begin{proof}
Define $Z_j = (y_j - \cA_I(j) )^2 - (y_j - \theta_{j})^2 - (\cA_I(j) - \theta_j)^2$.

Condition on the event $\mathcal{V}$ that $|\epsilon_t| \le \sigma \sqrt{2 \log (4n/\delta)}. \forall t \in [n]$ which happens with probability atleast $1-\delta/2$ by Lemma~\ref{lem:ub}. By Lemma~\ref{lem:selfbound}, we have $\{Z_j\}_{j \in \cT(I)}$ is a martingale difference sequence and $|Z_j| \le 16(B+\tilde \sigma)^2 = R_\sigma$. Note that once we condition on the filtration $\cF_j$, there is no randomness remaining in the terms $(\cA_I(j) - \theta_j)^2$ and $(\hat y_j - \theta_j)^2$. Hence $\mathbb{E}_{j}[(\cA_I(j) - \theta_j)^2|\mathcal{V}] = (\cA_I(j) - \theta_j)^2$ and $\mathbb{E}_{j}[(\hat y_j - \theta_j)^2|\mathcal{V}] = (\hat y_j - \theta_j)^2$. Using Lemma~\ref{lem:freedman} and taking $\eta = 1/R_\sigma$ we get,
\begin{align}
    \sum_{j \in \cT(I)} (y_j - \cA_I(j) )^2 - (y_j - \theta_{j})^2
    &\le \sum_{j \in \cT(I)} (e-1)(\cA_I(j) - \theta_j)^2 + R_\sigma^2 \log(4n \log n/\delta),
\end{align}
with probability atleast $1-\delta/(4n \log n)$ for a fixed expert $A_I$. Taking a union bound across all $O(n \log n)$ experts in $\cE$ leads to,
\begin{align}
    \mathbb{P}\left( \sum_{j \in \cT(I)} (y_j - \cA_I(j) )^2 - (y_j - \theta_{j})^2 \ge \sum_{j \in \cT(I)} (e-1) (\cA_I(j) - \theta_j)^2 + R_\sigma^2 \log(2n \log n/\delta )|\mathcal{V}\right) \le \delta/4,
\end{align}
for any expert $\cA_I$.

By similar arguments on the martingale difference sequence $(\hat{y}_j - \theta_j)^2 - (y_j - \hat{y}_j )^2 - (y_j + \theta_{j})^2$, it can be shown that
\begin{align}
    \mathbb{P}\left( \sum_{j \in \cT(I)} (y_j - \hat{y}_j )^2 - (y_j - \theta_{j})^2 \le \sum_{j \in \cT(I)} (3-e)(\hat{y}_j - \theta_j)^2 - R_\sigma^2 \log(2n \log n/\delta)|\mathcal{V}\right) \le \delta/4,
\end{align}
for any interval $I \in \cI|_{[n]}$. Taking union bound across the previous two bad events and multiplying the probability of noise boundedness event $\mathcal{V}$ leads to the lemma.
\end{proof}

\lemmstrongadapt*
\begin{proof}
Condition on the event $\mathcal{V}$. Then the losses $f_t(x) = (y_t - x)^2$ are $\frac{1}{4(B+\sigma \sqrt{\log( 2n / \delta)})^2} := \eta$ exp-concave \citep{Haussler1998mixloss,BianchiBook2006}. Since we pass $\eta \cdot f_t(x)$ as losses to SAA in \ALG{}, Lemma \ref{lem:saa} gives
\begin{align}
    \sum_{t \in \cT(I)} -\log\left(\sum_{J \in A_t} w_{t,J}e^{-\eta f_t(\cA_J(t))}\right) - \eta f_t(\cA_I(t))
    &\le \log (n \log n). \label{eq:saareg}
\end{align}
By $\eta$ exp-concavity of $f_t(x)$, we have
\begin{align}
    -\log\left(\sum_{J \in A_t} w_{t,J}e^{-\eta f_t(\cA_J(t)))}\right)
    &\ge \eta f_t\left(\sum_{J \in A_t} w_{t,J} \cA_I(t)\right),\\
    &= \eta f_t(\hat y_t). \label{eq:exp-concave} 
\end{align}
Combining \eqref{eq:saareg} and \eqref{eq:exp-concave} gives,
\begin{align}
    \sum_{t \in \cT(I)} f_t(\hat y_t) - f_t(\cA_I(t))
    &\le \frac{\log (n \log n)}{\eta},\\
    &\le \log (n \log n) R_\sigma.
\end{align}

So,
\begin{align}
    \sum_{t \in \cT(I)} (y_t - \hat y_t)^2 -(y_t - \theta_t)^2 
    &\le \sum_{t \in \cT(I)} (y_t - \cA_I(t))^2 -(y_t - \theta_t)^2 + \log (n \log n) R_\sigma,
\end{align}

Now invoking Lemma \eqref{lem:prob} followed by a trivial rearrangement completes the proof.

\end{proof}

\lemmbins*
\begin{proof}
We provide below a constructive proof. Consider the following scheme of partitioning $\mathbb{S}$.
\begin{enumerate}
    \item Set $\text{pings} = p(1), \text{TV} = 0, M = 1$.
    \item Start a partition from $x_1$.
    \item For i = 2 to m
    \begin{enumerate}
        \item If $\text{TV} + |\theta^{(i)} - \theta^{(i-1)}| > \frac{B}{\sqrt{\text{pings} + p(i)}}$:
        \begin{enumerate}
            \item $\text{pings} = p(i), \text{TV}=0$ // start a new bin (partition) from position $x_i$.
            \item $M = M + 1$ // increase the bin counter
        \end{enumerate}
        \item Else:
        \begin{enumerate}
            \item $\text{pings} = \text{pings} + p(i), \text{TV} = \text{TV} + |\theta^{(i)} - \theta^{(i-1)}|$
        \end{enumerate}
    \end{enumerate}
\end{enumerate}

Statements 1 and 2 of the Lemma trivially follows from the strategy. Next, we provide an upper bound on number of bins $M$ spawned by the above scheme. Let $[x_1,x_{r_1}], [x_{r_1+1}, x_{r_2}], \ldots, [x_{r_{M-1},x_{r_M}}]$ be the partition of $\mathbb{S}$ discovered by the above scheme.

Define the quantity $\text{TV}_1 := \sum_{i=1}^{r_1} |\theta^{(i)} - \theta^{(i+1)}|$ associated with bin 1. Similarly define $\text{TV}_2,\ldots,\text{TV}_{M-1}$ for other bins.

Define $N(1) = \sum_{i=1}^{r_1+1} p(i)$. Similarly define $N(2),\ldots N(M-1)$. It is immediate that $\sum_{i=1}^{M-1} N(i) \le 2n$.

We have,
\begin{align}
    C_n
    &\ge \sum_{i=1}^{M-1} \text{TV}_i,\\
    &\ge_{(1)} \sum_{i=1}^{M-1} \frac{B}{\sqrt{N(i)}},\\
    &\ge_{(2)} \frac{(M-1)^{3/2} \cdot B}{\sqrt{2n}},
\end{align}
where (1) follows from step 3(a) of the partitioning scheme and (2) is due to convexity of $1/\sqrt{x}, x > 0$ and applying Jensen's inequality. Rearranging and noting that $M-1 \ge M/2$, when $M >1$, we obtain
\begin{align}
    M
    &\le 3 n^{1/3}C_n^{2/3}B^{-2/3}.
\end{align}

Note that when $C_n = 0$, $M$ will remain 1 as a result of the partitioning scheme.

\end{proof}

\lemmexp*
\begin{proof}
Let $q(t) = \sum_{s=1}^{t-1} \bs 1\{i_s \in I\}$. Assume $q(t) > 0$. Fix a particular expert $\cA_I$ and a time $t$. Since $y_t \sim N(\theta_t,\sigma^2)$ by gaussian tail inequality we have,
\begin{align}
    \mathbb{P}\left( \left| \frac{\sum_{s=1}^{t-1} (y_s-\theta_s) \bs 1\{i_s \in I\}}{\sum_{s=1}^{t-1} \bs 1\{i_s \in I\}} \right| \ge \frac{\sigma}{\sqrt{q(t)}}\sqrt{\log\left(\frac{2n^3 \log n} {\delta}\right)}\right) \le \frac{\delta}{(n^3 \log n)}.
\end{align}
Applying a union bound across all time points and all experts implies that for any expert $\cA_I$ and $t \in \cT(I)$ with q(t) > 0,
\begin{align}
    \left | \cA_I(t) - \frac{\sum_{s=1}^{t-1} \theta_s \bs 1\{i_s \in I\}}{q(t)} \right|
    &\le \frac{\sigma}{\sqrt{q(t)}}\sqrt{\log\left(\frac{2n^3 \log n} {\delta}\right)}
\end{align}
with probability atleast $1-\delta$.

Now adding and subtracting $\theta_t$ inside the $|\cdot|$ on LHS and using $|a-b| \ge |a| - |b|$ yields,
\begin{align}
    |\cA_I(t) - \theta_t|
    &\le \left | \theta_t - \frac{\sum_{s=1}^{t-1} \theta_s \bs 1\{i_s \in I\}}{q(t)} \right| + \frac{\sigma}{\sqrt{q(t)}}\sqrt{\log\left(\frac{2n^3 \log n} {\delta}\right)}.
\end{align}

Hence,
\begin{align}
    \sum_{t \in \cT(I)} (\theta_t - \cA_I(t))^2
    &\le_{(a)} \sum_{t \in \cT(I)} 2\left( \theta_t - \frac{\sum_{s=1}^{t-1} \theta_s \bs 1\{i_s \in I\}}{q(t)}\right)^2 + 2\frac{\sigma^2}{q(t)}\log\left(\frac{2n^3 \log n} {\delta}\right)\\
    &\le \sum_{t \in \cT(I)} 2\left( \theta_t - \frac{\sum_{s=1}^{t-1} \theta_s \bs 1\{i_s \in I\}}{q(t)}\right)^2 + 2 \sigma^2 \log(|\cT(I)|)\log\left(\frac{2n^3 \log n} {\delta}\right), \label{eq:red1}
\end{align}
with probability atleast $1-\delta$. In (a) we used the relation $(a+b)^2 \le 2a^2 + 2b^2$.

Further we have,
\begin{align}
    \sum_{t \in \cT(I)} 2\left( \theta_t - \frac{\sum_{s=1}^{t-1} \theta_s \bs 1\{i_s \in I\}}{q(t)}\right)^2
    &\le 2 V(\ubar{x}, \bar{x})^2 | \cT(I)|. \label{eq:red2}
\end{align}

Combining \eqref{eq:red1} and \eqref{eq:red2} completes the proof.

\end{proof}

\subsection{Proof of the main result: Theorem~\ref{thm:main}}\label{part3}
\begin{proof}
Throughout the proof we carry forward all notations used in Lemmas \ref{lem:bins} and \ref{lem:expert-concent}.

We will apply Corollary \ref{cor:bins} to the partition in Lemma~\ref{lem:bins}. Take a specific partition $[x_i,x_j] \in \mathcal{P}$ with $j \neq m$. Consider a set of indices $F = \{k_i,k_{i}+1,\ldots,k_j \}$ of consecutive natural numbers between $k_i$ and $k_j$. By Proposition \ref{prop:gc} $F$ can be covered using elements in $\cI|_{[n]}$. Let this cover be $\mathcal{U}$. For any  $I \in \mathcal{U}$, we have
\begin{align}
    \sum_{t \in T(I)} (\theta_t - \cA_I(t))^2
    &\le_{(a)} 2V(\ubar{x},\bar{x})^2 |\cT(I)| +  2\sigma^2 \log (2n^3 \log n/\delta) \log(|\cT(I)|)\\
    &\le 2V(\ubar{x},\bar{x})^2 |\cT(F)| +  2\sigma^2 \log (2n^3 \log n/\delta) \log(|\cT(I)|)\\
    &\le_{(b)} 2B^2 + 2\sigma^2 \log (2n^3 \log n/\delta) \log(n),
\end{align},
with probability atleast $1-\delta$. Step (a) is due to Lemma~\ref{lem:expert-concent} and (b) is due to statement 1 of Lemma~\ref{lem:bins}.

Using Lemma~\ref{lem:strong-adapt} and a union bound on the bad events in Lemmas \ref{lem:strong-adapt} and \ref{lem:expert-concent} yields,
\begin{align}
    \sum_{t \in \cT(I)} (\hat y_t - \theta_t)^2
    &\le \frac{e-1}{3-e} \left( 2B^2 + 2\sigma^2 \log (2n^3 \log n/\delta) \log(n) \right) + \lambda,    
\end{align}
with probability atleast $1-2\delta$ and $\lambda$ is as defined in Corollary \ref{cor:bins}. Due to the property of exponentially decaying lengths as stipulated by Proposition \ref{prop:gc}, there are only atmost $2 \log |F| \le 2 \log n$ intervals in $\mathcal{U}$. So,
\begin{align}
     \sum_{t \in \cT(F)} (\hat y_t - \theta_t)^2
     &\le 2 \log n \left( \frac{e-1}{3-e} \left( 2B^2 + 2\sigma^2 \log (2n^3 \log n/\delta) \log(n) \right) + \lambda \right).
\end{align}

Similar bound can be obtained for the last bin $[x_{r_{M-1}+1},x_{m}]$ in $\mathcal{P}$. There are two cases to consider. In case 1, we consider the scenario when $V(x_{r_{M-1}+1},x_{m})$ obeys relation 1 of Lemma~\ref{lem:bins}. Then the analysis is identical to the one presented above. In case 2, we consider the scenario when $V(x_{r_{M-1}+1},x_{m-1})$ obeys relation 2 of Lemma~\ref{lem:bins} while $V(x_{r_{M-1}+1},x_{m})$ doesn't. Then the error incurred within the interior $[x_{r_{M-1}+1},x_{m-1}]$ can be bounded as before. To bound the error at last point, we only need to bound the error of expert that performs mean estimation of iid gaussians. It is well known that the cumulative squared error for this problem is atmost $\sigma^2 \log (n/\delta)$ with probability atleast $1-\delta$.

By Lemma~\ref{lem:bins}, $|\mathcal{P}| = \max\{3n^{1/3}C_n^{2/3}B^{-2/3},1\}$. Hence the total error summed across all partitions in $\mathcal{P}$ becomes,
\begin{align}
    \begin{split}
    \sum_{t=1}^{n} (\hat y_t - \theta_t)^2
    \le{}&  2 \log n \left( \frac{e-1}{3-e} \left( 4n^{1/3}C_n^{2/3}B^{4/3} + 4\sigma^2 \log (2n^3 \log n/\delta) \log(n)n^{1/3}C_n^{2/3}B^{-2/3} \right)\right)\\
    &+ 4 \log(n)  \frac{e-1}{3-e} \lambda n^{1/3}C_n^{2/3}B^{-2/3}\\
    &+  2 \log(n) \left( \frac{e-1}{3-e} \left( 2B^2 + 2\sigma^2 \log (2n^3 \log n/\delta) \log(n) \right) + \lambda \right) + \sigma^2 \log (n/\delta), \label{eq:fin}\\
    &= \tilde{O}(n^{1/3}C_n^{2/3}),
    \end{split}
\end{align}
with probability atleast $1-2\delta$. A change of variables from $2\delta \rightarrow \delta$ completes the proof. As a closing note, we remark that the aggressive dependence of $B$ in \eqref{eq:fin} on cases when $B$ is too small can be dampened by using a threshold of $\frac{1}{\sqrt{\text{pings} + p(i)}}$ in the partition scheme presented in proof of Lemma~\ref{lem:bins}.
\end{proof}

\section{Excluded details in Experimental section} \label{app:exp}
\begin{figure}[tbh]
\centering
\includegraphics[width=0.8\textwidth]{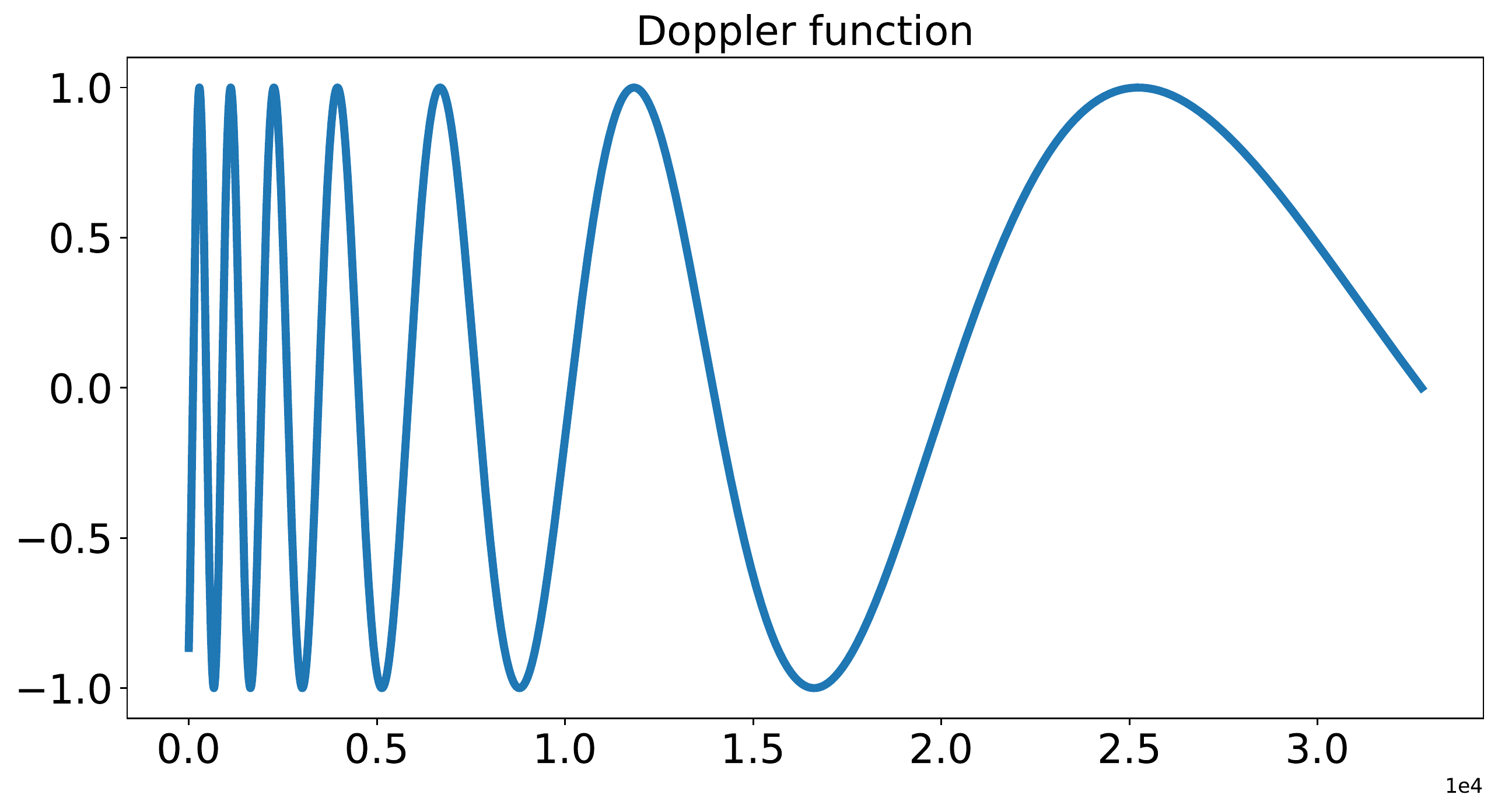}
\caption{\emph{Doppler function, TV = 27}}
\label{fig:doppler}
\end{figure}

\begin{figure}[tbh]
\centering
\includegraphics[width=0.8\textwidth]{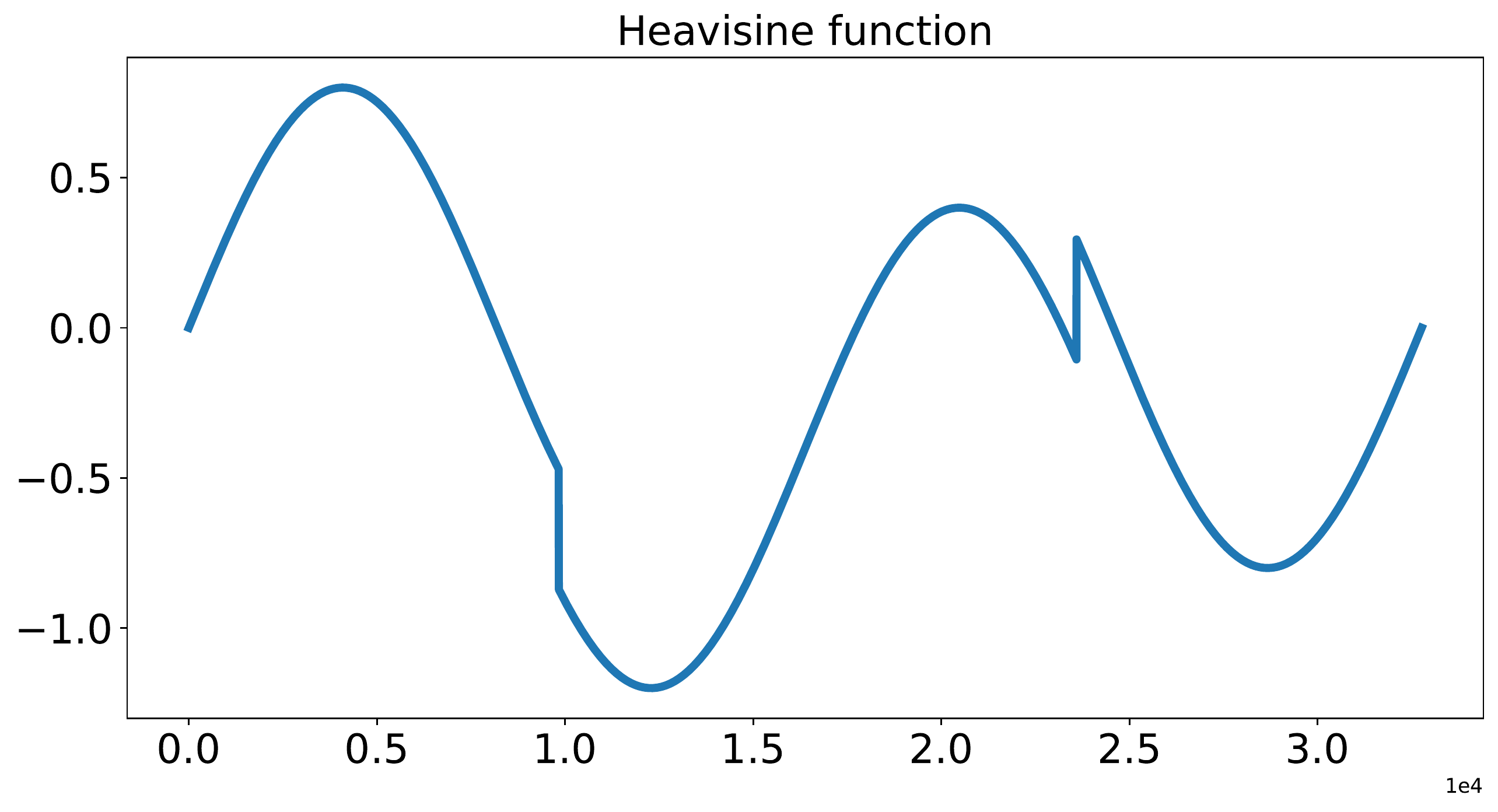}
\caption{\emph{Heavisine function, TV = 7.2}}
\label{fig:heavi}
\end{figure}

\textbf{Waveforms.} The waveforms shown in Fig. \ref{fig:doppler} and \ref{fig:heavi} are borrowed from \citep{donoho1994ideal}. Note that both functions exhibit spatially inhomogeneous smoothness behaviour.

\begin{figure}[H]
\centering
\includegraphics[width=0.8\textwidth]{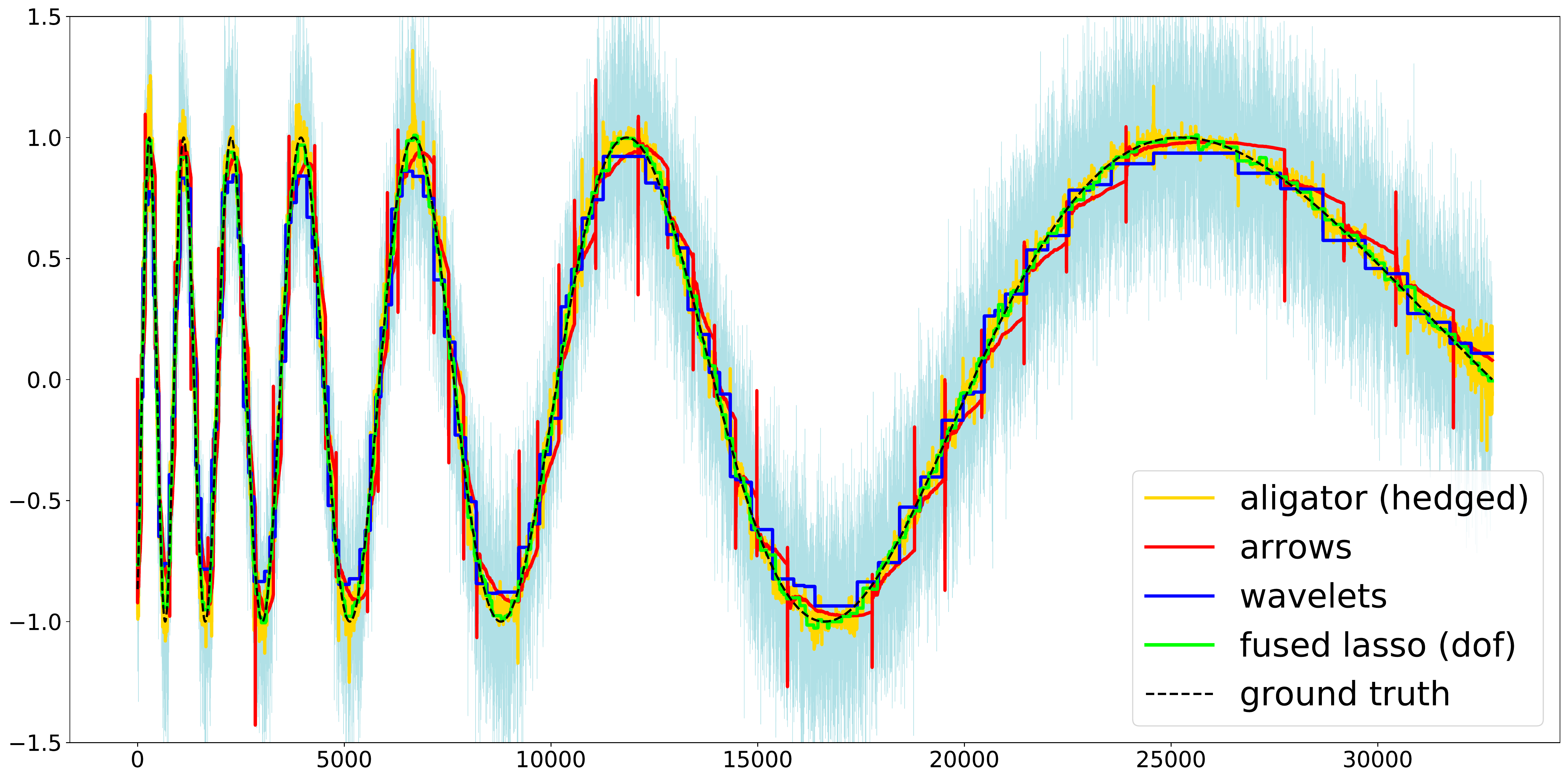}
\caption{\emph{Fitted signals for Doppler function with noise level $\sigma = 0.35$}}
\end{figure}

\begin{figure}[H]
\minipage{0.32\textwidth}
  \includegraphics[width=\linewidth]{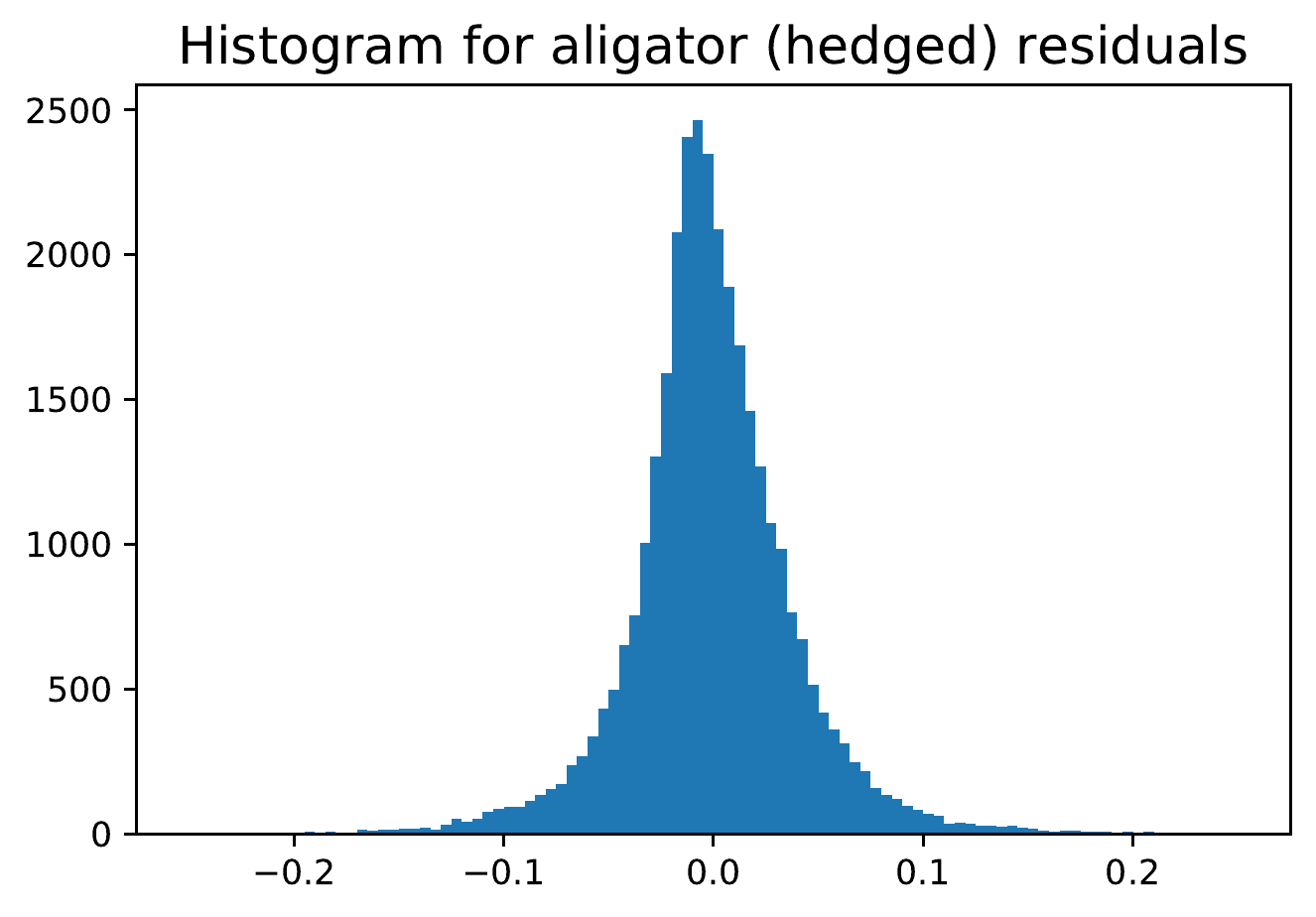}
\endminipage\hfill
\minipage{0.32\textwidth}
  \includegraphics[width=\linewidth]{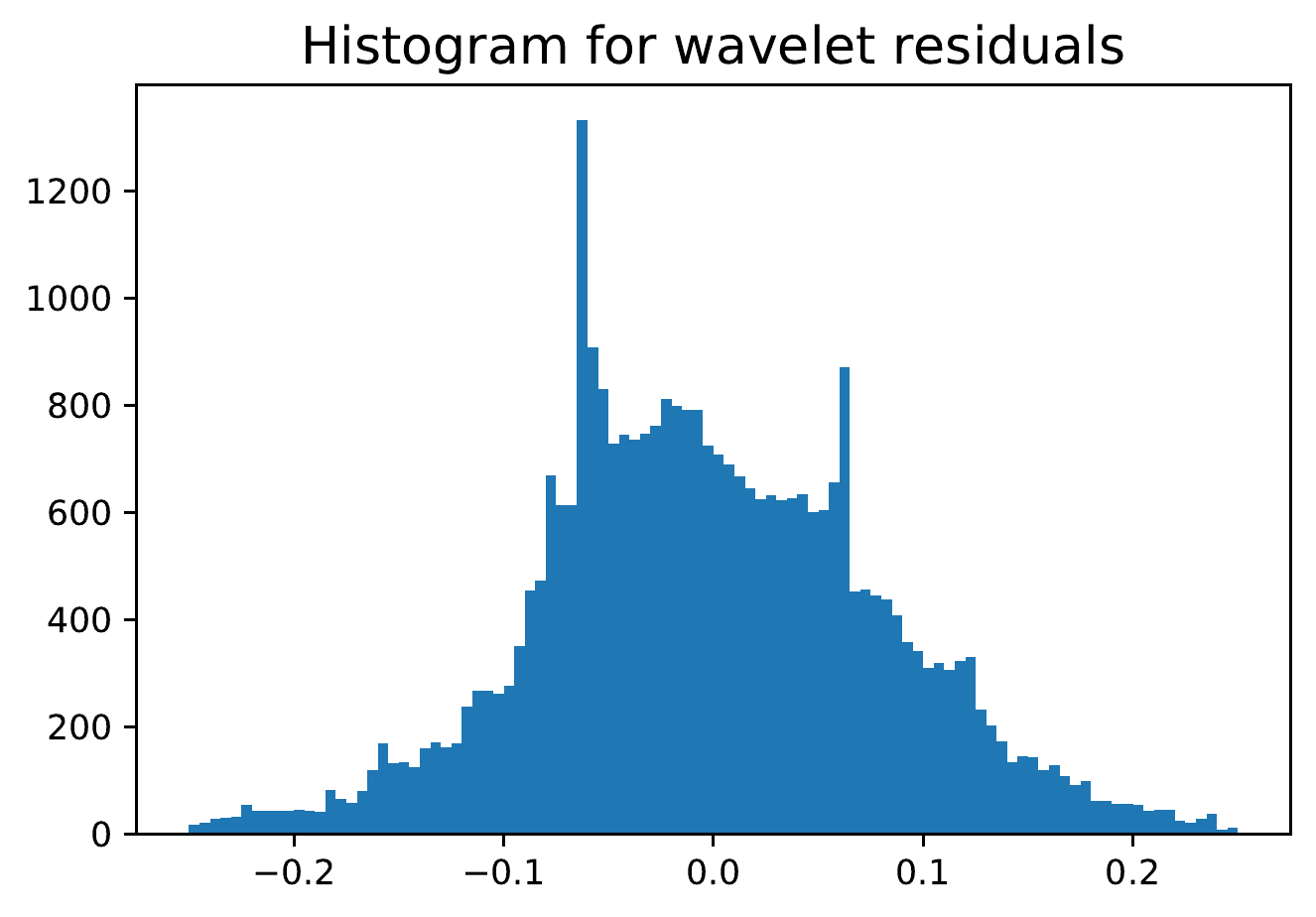}
\endminipage\hfill
\minipage{0.32\textwidth}%
  \includegraphics[width=\linewidth]{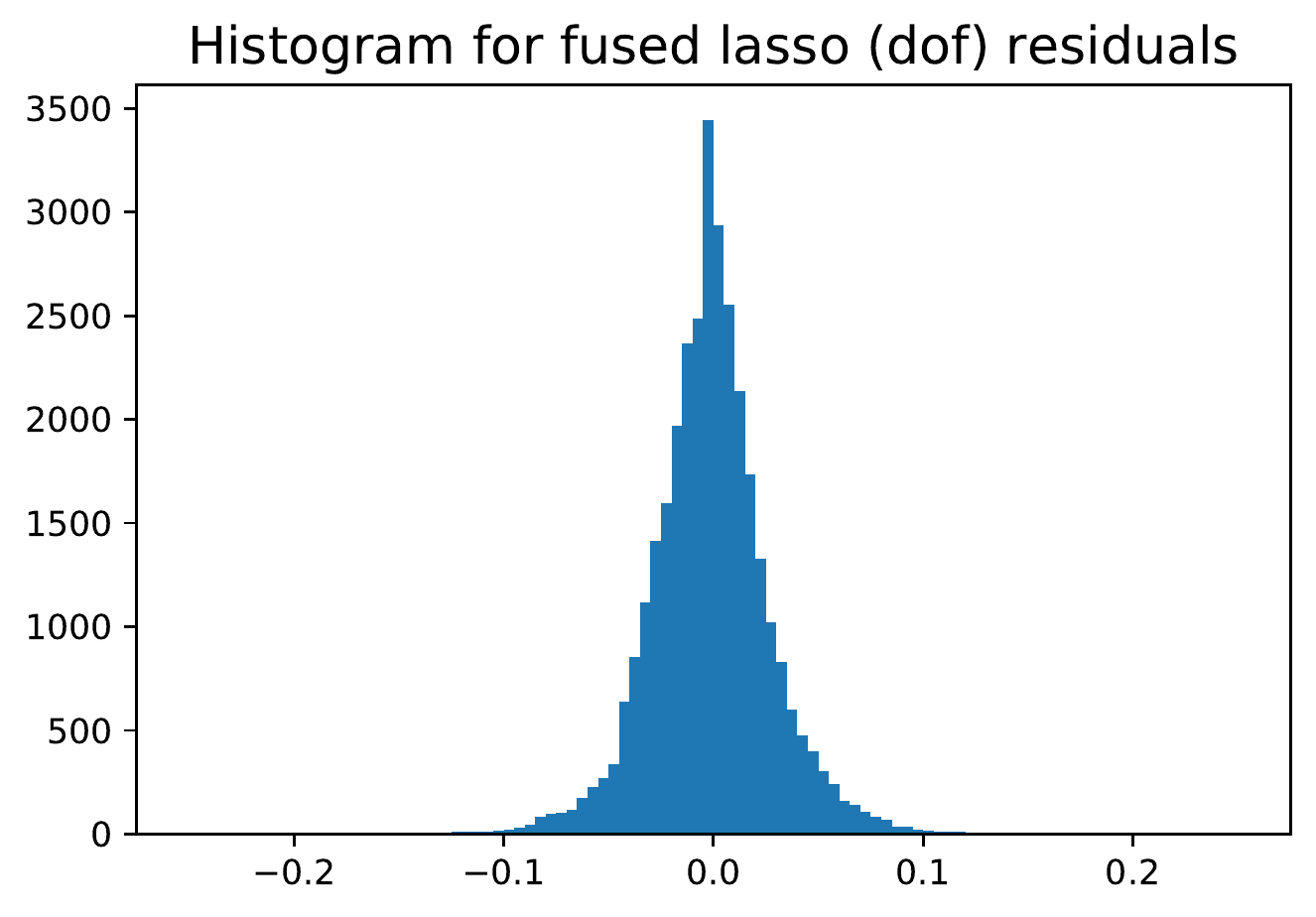}
\endminipage
\caption{\emph{Histogram of residuals for various algorithms when run on Doppler function with noise level $\sigma=0.35$. Note that they are residuals w.r.t to ground truth. \ALG{} incurs lower bias than wavelets. The bias incurred by dof fused lasso is roughly comparable to \ALG{} while former is more compute intensive. }}
\end{figure}

\begin{figure}[H]
\centering
\includegraphics[width=0.8\textwidth]{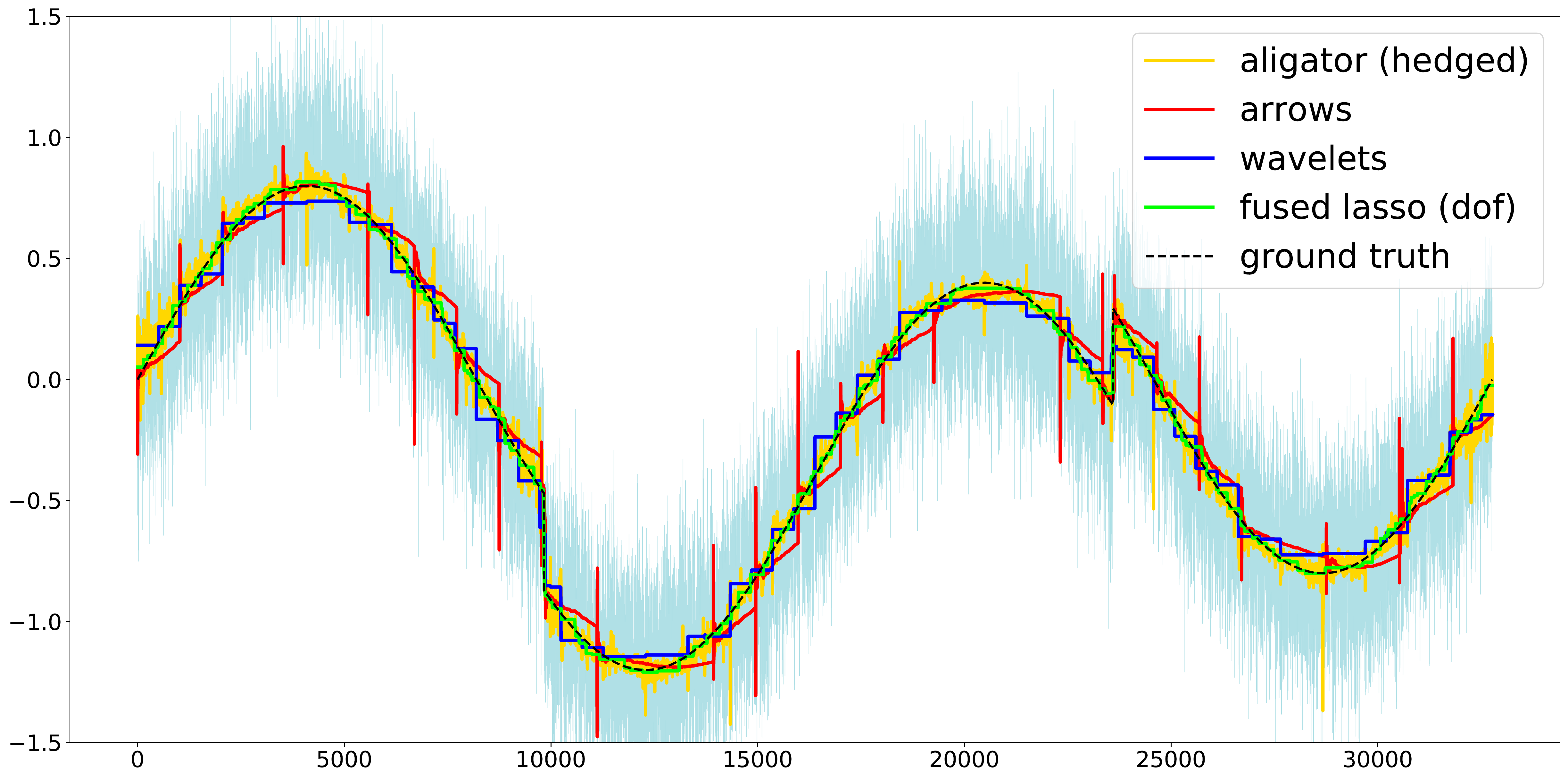}
\caption{\emph{Fitted signals for Heavisine function with noise level $\sigma = 0.35$}}
\end{figure}

\begin{figure}[H]
\minipage{0.32\textwidth}
  \includegraphics[width=\linewidth]{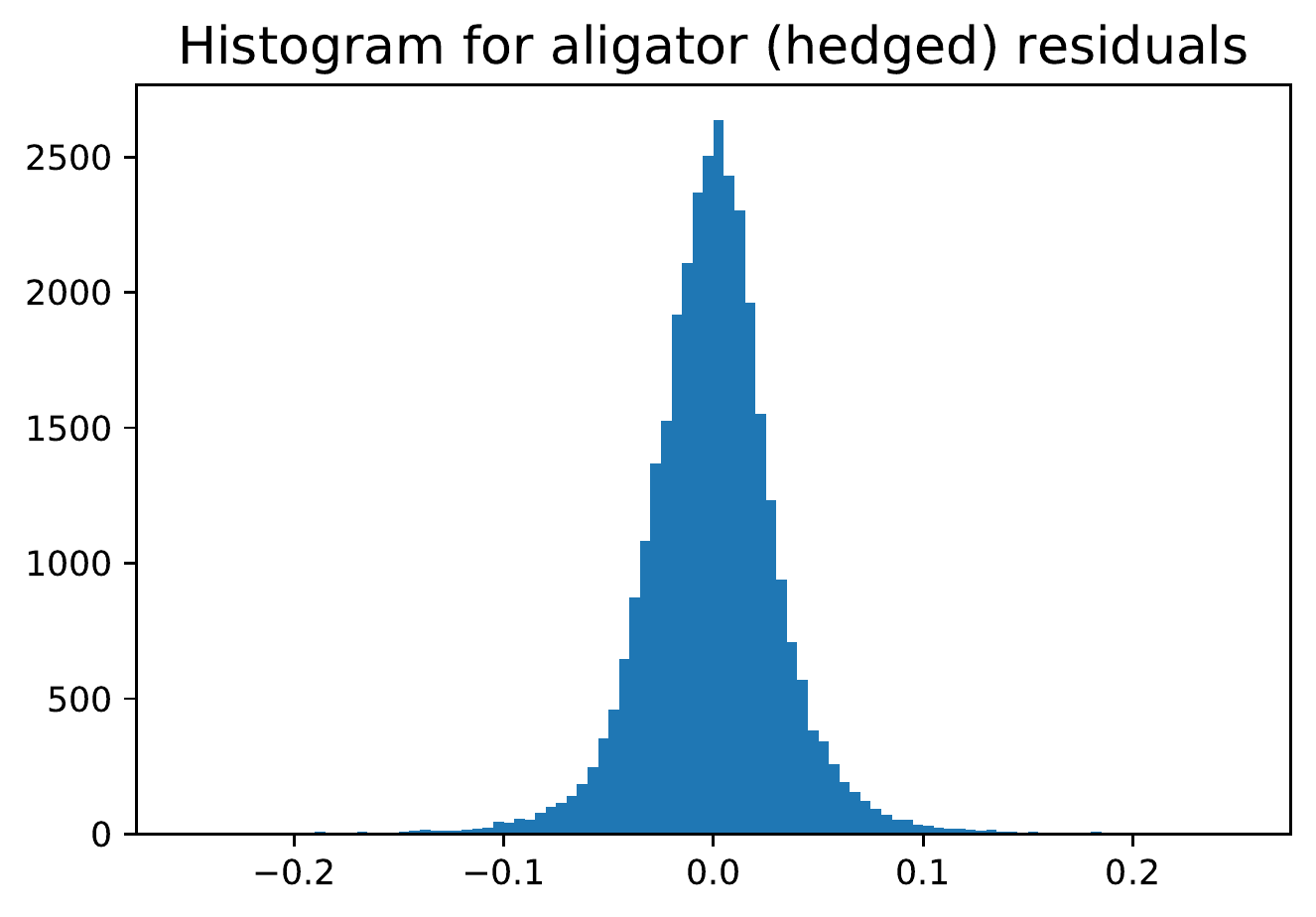}
\endminipage\hfill
\minipage{0.32\textwidth}
  \includegraphics[width=\linewidth]{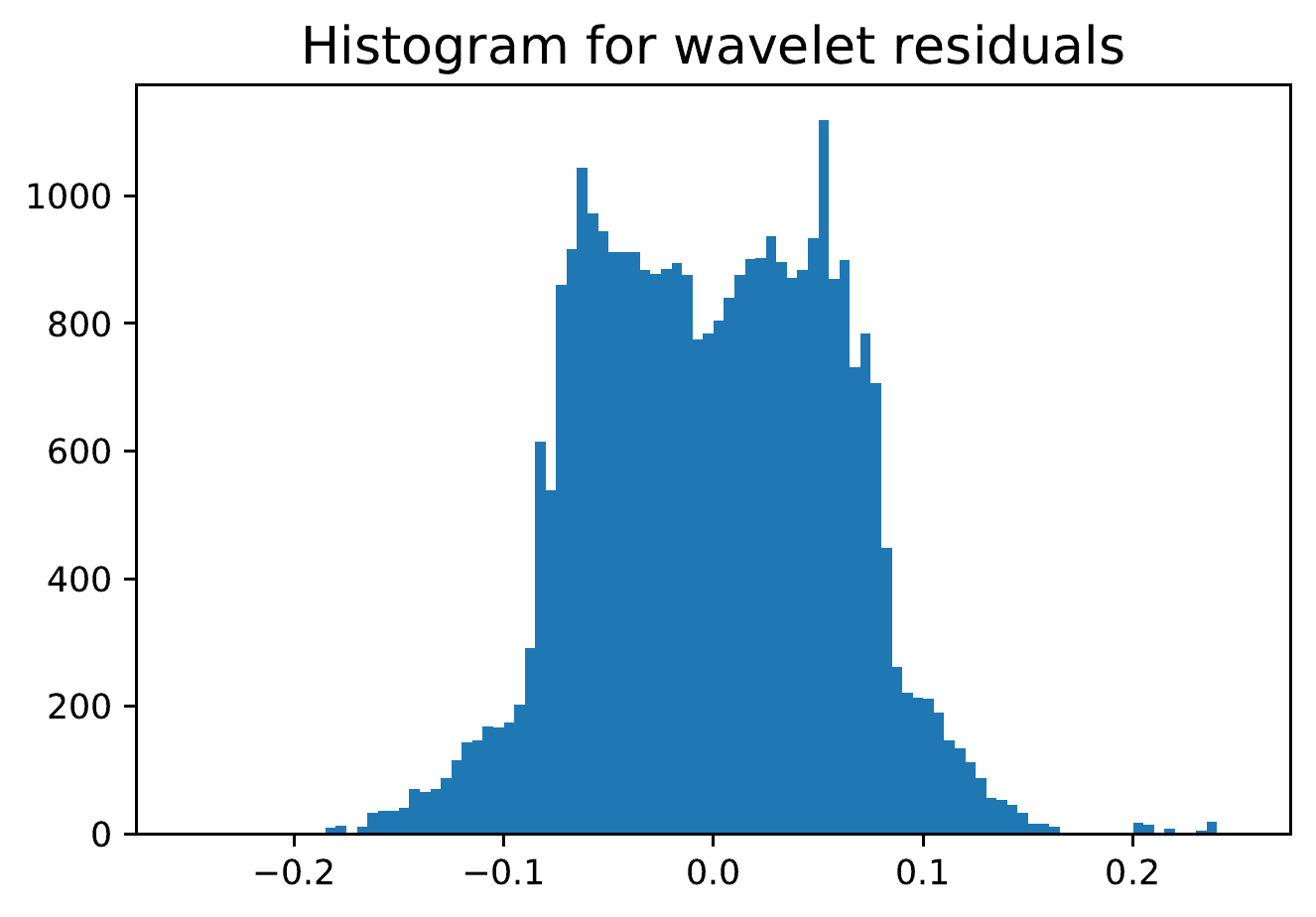}
\endminipage\hfill
\minipage{0.32\textwidth}%
  \includegraphics[width=\linewidth]{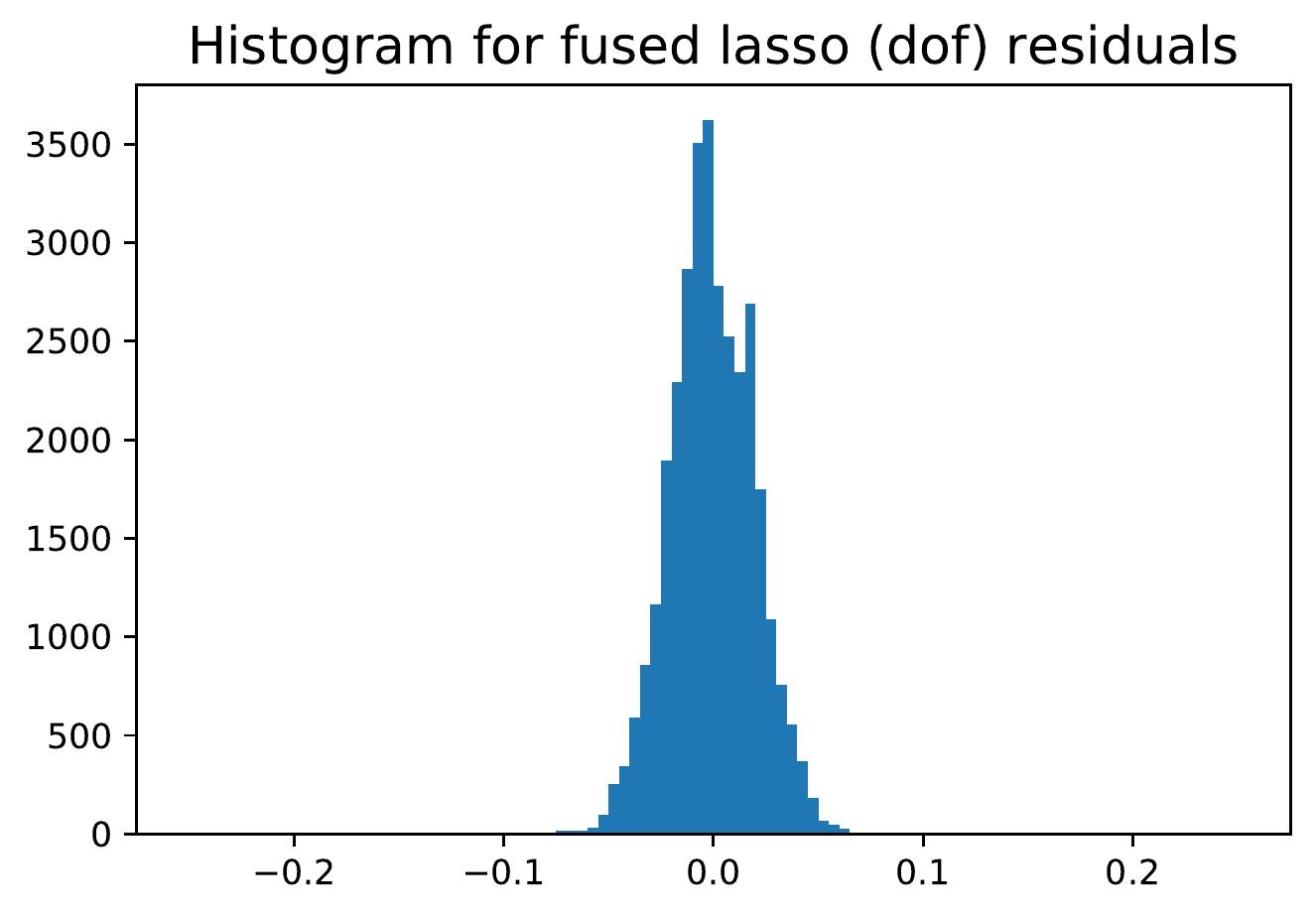}
\endminipage
\caption{\emph{Histogram of residuals for various algorithms when run on Heavisine function with noise level $\sigma=0.35$. Note that they are residuals w.r.t to ground truth. \ALG{} incurs lower bias than wavelets. The bias incurred by dof fused lasso is roughly comparable to \ALG{} while former is more compute intensive.}}
\end{figure}

\begin{figure}[H]
\minipage{0.48\textwidth}
  \includegraphics[width=\linewidth]{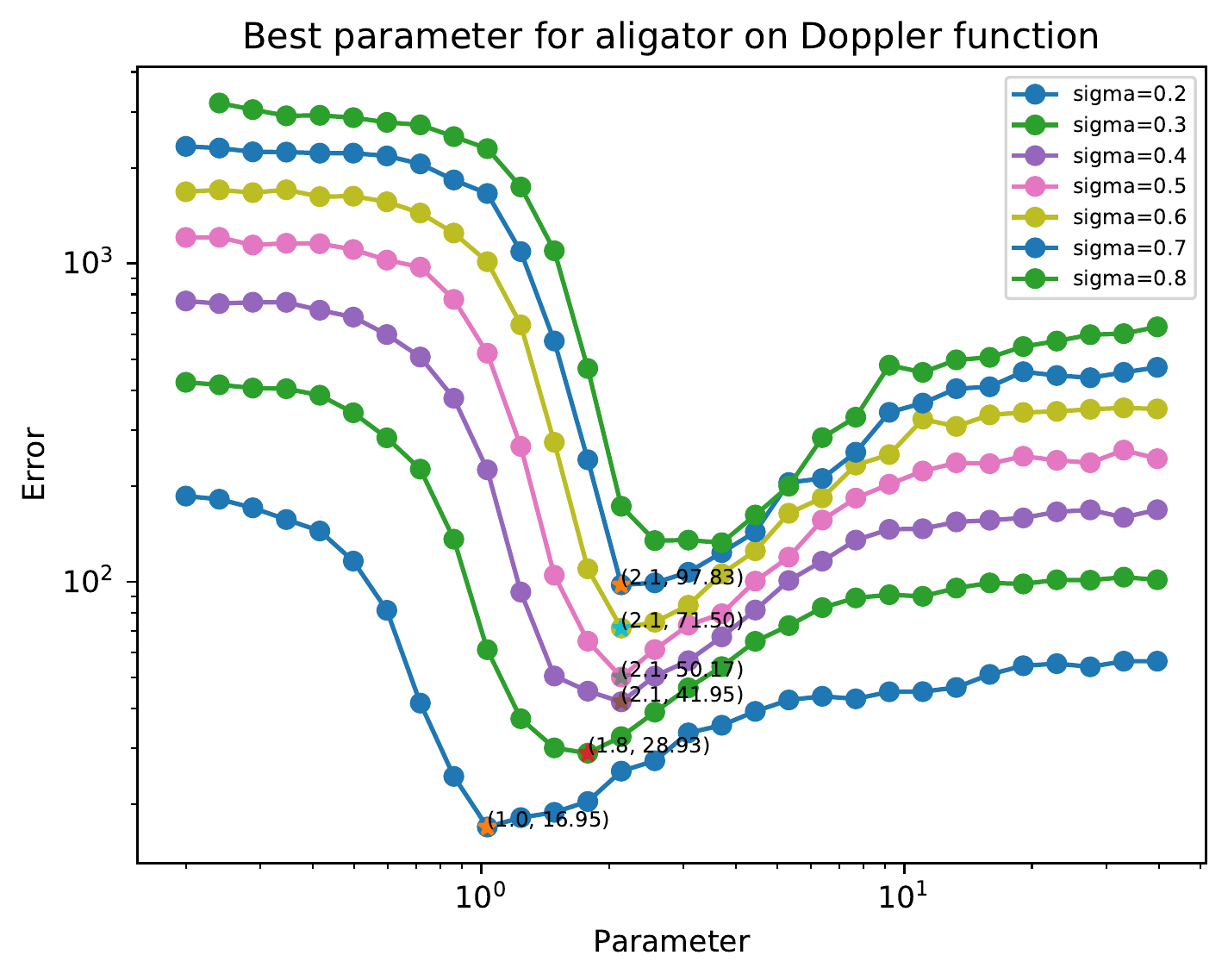}
\endminipage\hfill
\minipage{0.48\textwidth}
  \includegraphics[width=\linewidth]{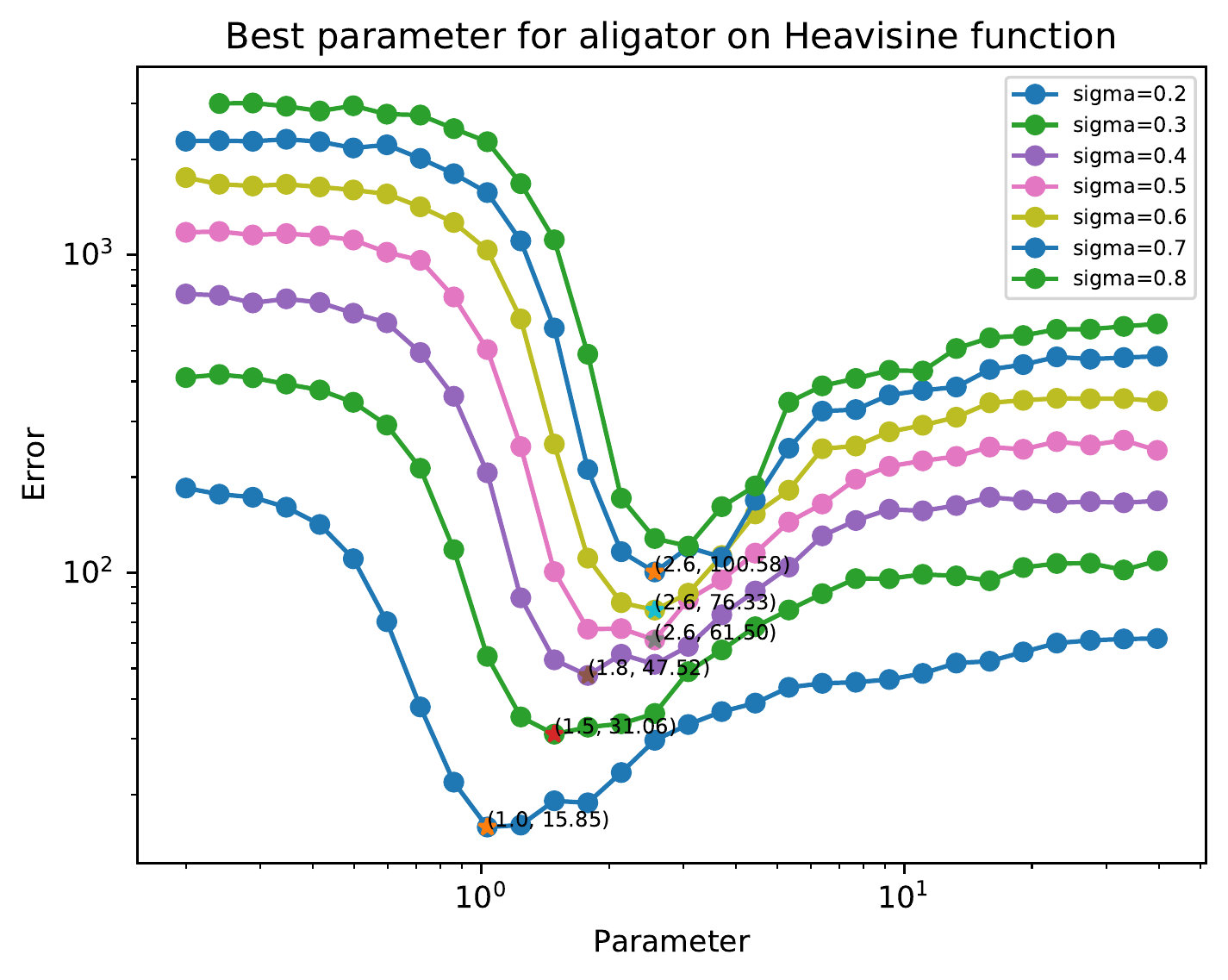}
\endminipage
\caption{\emph{Hyper-parameter search for learning rate in \ALG{} \textit{(heuristics)}.}}
\label{fig:compareloss}
\end{figure}
\textbf{Hyper-parameter search.} Initially we used a grid search on an exponential grid to realize that the optimal $\lambda$ across all experiments fall within the range [0.125, 8]. Then we used a fine-tuned grid $[0.125,0.25,0.5,0.75,1,1.5,2,2.5,3,3.5,4,4.5,5,5.5,6,6.5,7,7.5,8,10,12,14,16]$ to search for the final hyper parameter value. For \ALG{} \textit{(heuristics)}, we searched for different noise levels in order to find best learning rate. We set search method as $ \text{Loss}/ (para * (\sigma ^ 2 + \sigma^2 / m))$. As Fig. \ref{fig:compareloss} shows,  $para = 2$ is found to provide good results across all signals we consider.

\textbf{Padding for wavelets.} For ``wavelet'' estimator in Fig. \ref{fig:experiment}, when data length is not a power of 2, we used the reflect padding mode in \citep{pywavelets}, though the results are similar for other padding schemes.

\begin{figure}[H]   
  \centering
  \stackunder{\hspace*{-0.5cm}\includegraphics[width=0.75\textwidth,height=0.5\textheight,keepaspectratio]{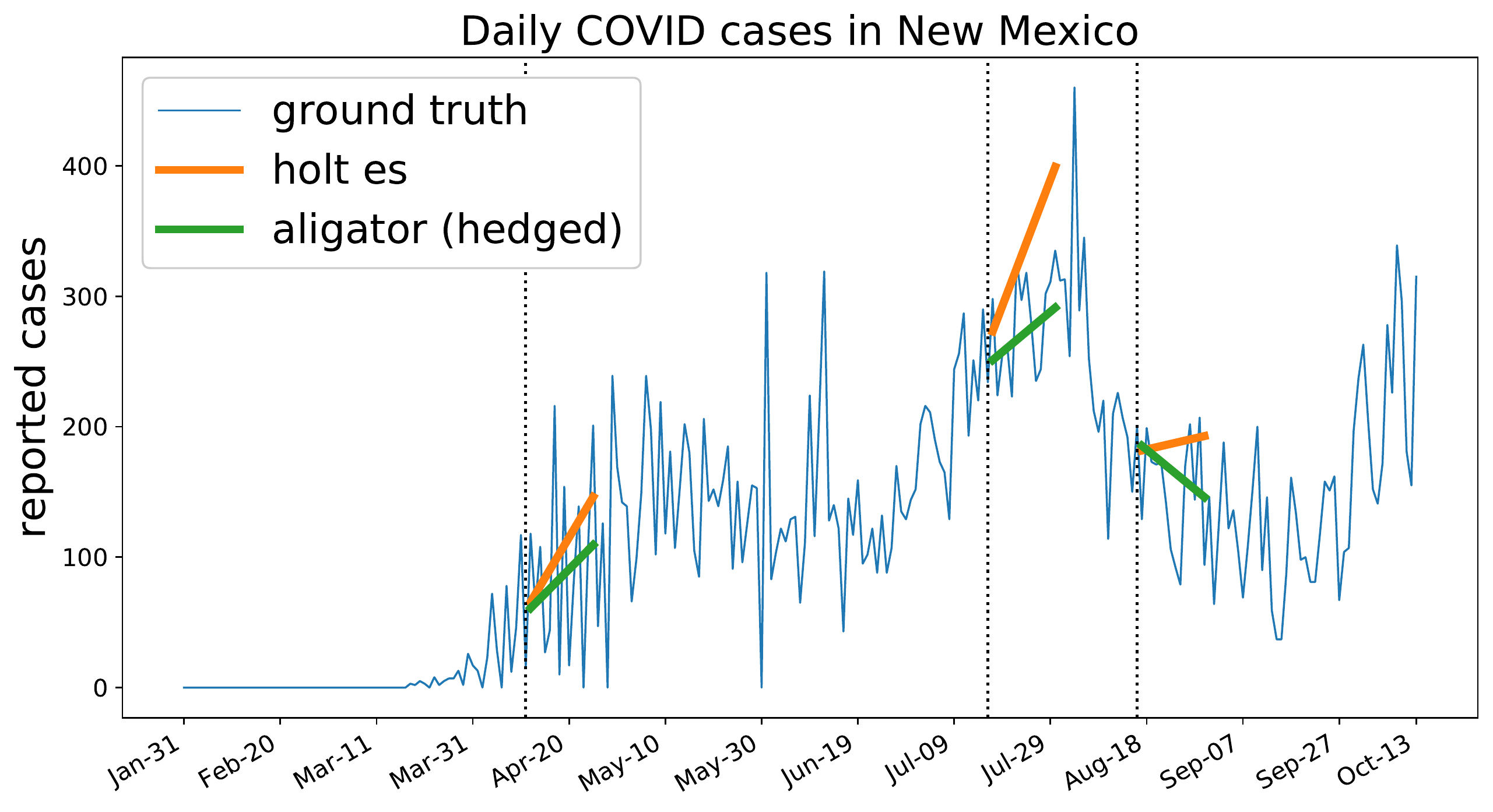}}{}
  \caption{\emph{A demo on forecasting COVID cases based on real world data. We display the two weeks forecasts of hedged \ALG{} and Holt ES, starting from the time points identified by the dotted lines. Both the algorithms are trained on a 2 month data prior to each dotted line. We see that hedged \ALG{} detects changes in trends more quickly than Holt ES. Further, hedged \ALG{} attains a 12\% reduction in the average RMSE from that of Holt ES (see Table \ref{tab:states}).}}
  \label{fig:new mexico}
\end{figure}

\textbf{Experiments on Real Data.} We follow the experimental setup described in Section \ref{sec:exp}. A qualitative comparison of the forecasts for the state of New Mexico, USA is illustrated in Fig. \ref{fig:new mexico}. The average RMSE of \ALG{} and Holt ES for all states in USA is reported in Table \ref{tab:states}.

\begin{table}[]
\begin{center}
\begin{tabular}{|l|c|c|c|}
\hline
\textbf{State} & \begin{tabular}[c]{@{}l@{}} \textbf{RMSE}\\ \textbf{Aligator}\end{tabular} & \begin{tabular}[c]{@{}l@{}} \textbf{RMSE}\\ \textbf{Holt ES}\end{tabular} & \textbf{\% improvement} \\ \hline
New Jersey & 411.87 & 546.89 & 24.69 \\ \hline
Ohio & 216.24 & 280.24 & 22.84 \\ \hline
Florida & 1330.33 & 1671.23 & 20.4 \\ \hline
Alabama & 290.71 & 362.13 & 19.72 \\ \hline
New York & 876.35 & 1054.2 & 16.87 \\ \hline
Rhode Island & 85.11 & 98.23 & 13.35 \\ \hline
Vermont & 7.59 & 8.7 & 12.76 \\ \hline
Kansas & 142.17 & 162.16 & 12.33 \\ \hline
New Mexico & 57.88 & 65.99 & 12.29 \\ \hline
Connecticut & 206.79 & 235.6 & 12.23 \\ \hline
California & 1456.48 & 1650.25 & 11.74 \\ \hline
Pennsylvania & 258.21 & 290.6 & 11.14 \\ \hline
Kentucky & 145.61 & 163.59 & 10.99 \\ \hline
New Hampshire & 25.16 & 27.99 & 10.1 \\ \hline
Minnesota & 161.41 & 179.12 & 9.89 \\ \hline
Michigan & 315.86 & 350.24 & 9.82 \\ \hline
Hawaii & 30.24 & 33.18 & 8.86 \\ \hline
Texas & 1510.42 & 1650.73 & 8.5 \\ \hline
South Dakota & 56.83 & 61.8 & 8.04 \\ \hline
Utah & 118.97 & 128.96 & 7.74 \\ \hline
Alaska & 17.54 & 18.96 & 7.52 \\ \hline
Washington & 188.8 & 202.74 & 6.88 \\ \hline
North Carolina & 265.74 & 284.47 & 6.58 \\ \hline
Nebraska & 98.49 & 105.41 & 6.56 \\ \hline
Montana & 28.31 & 30.28 & 6.51 \\ \hline
Missouri & 224.51 & 239.9 & 6.42 \\ \hline
Iowa & 205.77 & 219.28 & 6.16 \\ \hline
District of Columbia & 33.58 & 35.74 & 6.04 \\ \hline
Virginia & 194.29 & 206.44 & 5.89 \\ \hline
Nevada & 159.88 & 168.92 & 5.35 \\ \hline
Wyoming & 16.43 & 17.25 & 4.73 \\ \hline
Georgia & 493.93 & 518.27 & 4.7 \\ \hline
Oregon & 55.48 & 58.21 & 4.68 \\ \hline
Louisiana & 562.89 & 590.49 & 4.67 \\ \hline
Maryland & 209.95 & 218.22 & 3.79 \\ \hline
Illinois & 475.49 & 492.09 & 3.37 \\ \hline
West Virginia & 37.34 & 38.63 & 3.33 \\ \hline
Delaware & 64.1 & 66.26 & 3.26 \\ \hline
Tennessee & 384.55 & 396.95 & 3.12 \\ \hline
Arizona & 481.91 & 493.73 & 2.39 \\ \hline
South Carolina & 271.87 & 277.42 & 2.0 \\ \hline
Idaho & 93.83 & 95.44 & 1.68 \\ \hline
Colorado & 142.58 & 144.53 & 1.35 \\ \hline
Mississippi & 206.67 & 209.11 & 1.16 \\ \hline
Arkansas & 164.83 & 164.88 & 0.03 \\ \hline
Massachusetts & 302.79 & 301.8 & \red{-0.32} \\ \hline
Oklahoma & 151.82 & 146.65 & \red{-3.41} \\ \hline
Indiana & 185.1 & 178.2 & \red{-3.73} \\ \hline
North Dakota & 42.14 & 40.49 & \red{-3.92} \\ \hline
Wisconsin & 219.04 & 203.37 & \red{-7.15} \\ \hline
Maine & 14.59 & 13.37 & \red{-8.36} \\ \hline\end{tabular}
\caption{Average RMSE across all states in USA. The experimental setup and computation of error metrics are as described in Section \ref{sec:exp}. The \% improvement tab is computed as follows. Let $x_1$ and $x_2$ be the RMSE of \ALG{} and Holt ES respectively. Then $\text{\% improvement} = (x_2 - x_1)/\max\{x_1,x_2\}$.} \label{tab:states}
\end{center}
\end{table}

\end{document}